\documentclass[journal]{IEEEtran}
\usepackage{cite}
\usepackage{amsmath}
\usepackage{algorithmic}
\usepackage{array}
\usepackage{algorithm}
\usepackage{algorithmic}
\usepackage{bm}
\usepackage{booktabs}
\usepackage{amsmath}
\usepackage{amssymb}
\usepackage{graphicx}
\usepackage{mathtools}
\usepackage{multirow}
\usepackage{pifont} 
\usepackage[table]{xcolor}
\definecolor{bestblue}{RGB}{205,227,255}
\definecolor{secondblue}{RGB}{243,249,255}
\definecolor{abred1}{RGB}{244,156,156}  
\definecolor{abred2}{RGB}{247,186,186}   
\definecolor{abred3}{RGB}{250,214,214}  
\definecolor{abred4}{RGB}{252,234,234}  
\definecolor{abred5}{RGB}{254,246,246}  

\usepackage{comment}

\begin{document}
\title{MODAL: Multi-Modal Object Re-ID via Model-Driven Sparse Decoupling and Text-Image Differential Filtering}
\author{
Chengbo~Huang,
Jun-Jie~Huang,
Long~Lan,
Tianrui~Liu,
Xueqiong~Li,
Yuanxi~Peng,
Xinwang~Liu,~\IEEEmembership{Senior Member,~IEEE}
and~Meng~Wang,~\IEEEmembership{Fellow,~IEEE}
\thanks{C. Huang, J.-J. Huang, L. Lan, T. Liu, X. Li, Y. Peng and X. Liu are with the College of Computer Science and Technology, National University of Defense Technology, Changsha 410073, China (e-mail: chengbohuang@nudt.edu.cn; jjhuang@nudt.edu.cn; long.lan@nudt.edu.cn; trliu@nudt.edu.cn; lixueqiong13@nudt.edu.cn; pyx@nudt.edu.cn; xinwangliu@nudt.edu.cn).}
\thanks{M. Wang is with the School of Computer Science and Information Engineering, Hefei University of Technology, Hefei 230002, China (e-mail: eric.mengwang@gmail.com).}
}

\maketitle

\begin{abstract}
Multi-modal object re-identification (Re-ID) aims to facilitate cross-camera object retrieval in complex environments by leveraging complementary information from visual (e.g., RGB, NIR, TIR) and textual modalities. 
However, existing approaches often lack principled feature disentanglement and coherent multi-modal integration, leading to entangled representations that introduce cross-modal conflicts, obscure discriminative cues, and suffer distribution shift under modality-missing conditions.
To tackle these challenges, we propose MODAL, a novel multi-modal object re-identification framework,
grounded in coupled sparse coding theory and differential suppression principles. 
A core component of MODAL is a Multi-modal Feature Sparse Decoupling module, developed in a model-driven deep unrolling manner based on multi-modal coupled sparse coding. It explicitly decomposes multi-modal features into uni-modal specific, bi-modal and tri-modal shared representations, thereby achieving more transparent and effective feature disentanglement. Benefiting from the principled feature disentanglement, MODAL naturally mitigates performance degradation in incomplete-modality scenarios via a Modality-Aware Subspace Activation that selectively activates only the consistently shared subspaces.
Moreover, we propose a Text-Image Differential Filtering module that leverages coarse-grained textual semantics to adaptively suppress task-irrelevant responses in the decoupled visual representations, thereby enhancing discriminative information. Extensive experiments on four datasets demonstrate that MODAL achieves state-of-the-art performance with superior transparency. 
\end{abstract}


\IEEEpeerreviewmaketitle

\section{Introduction}
\label{sec:intro}

\begin{figure}[t]
    \centering
    \includegraphics[width=1\columnwidth]{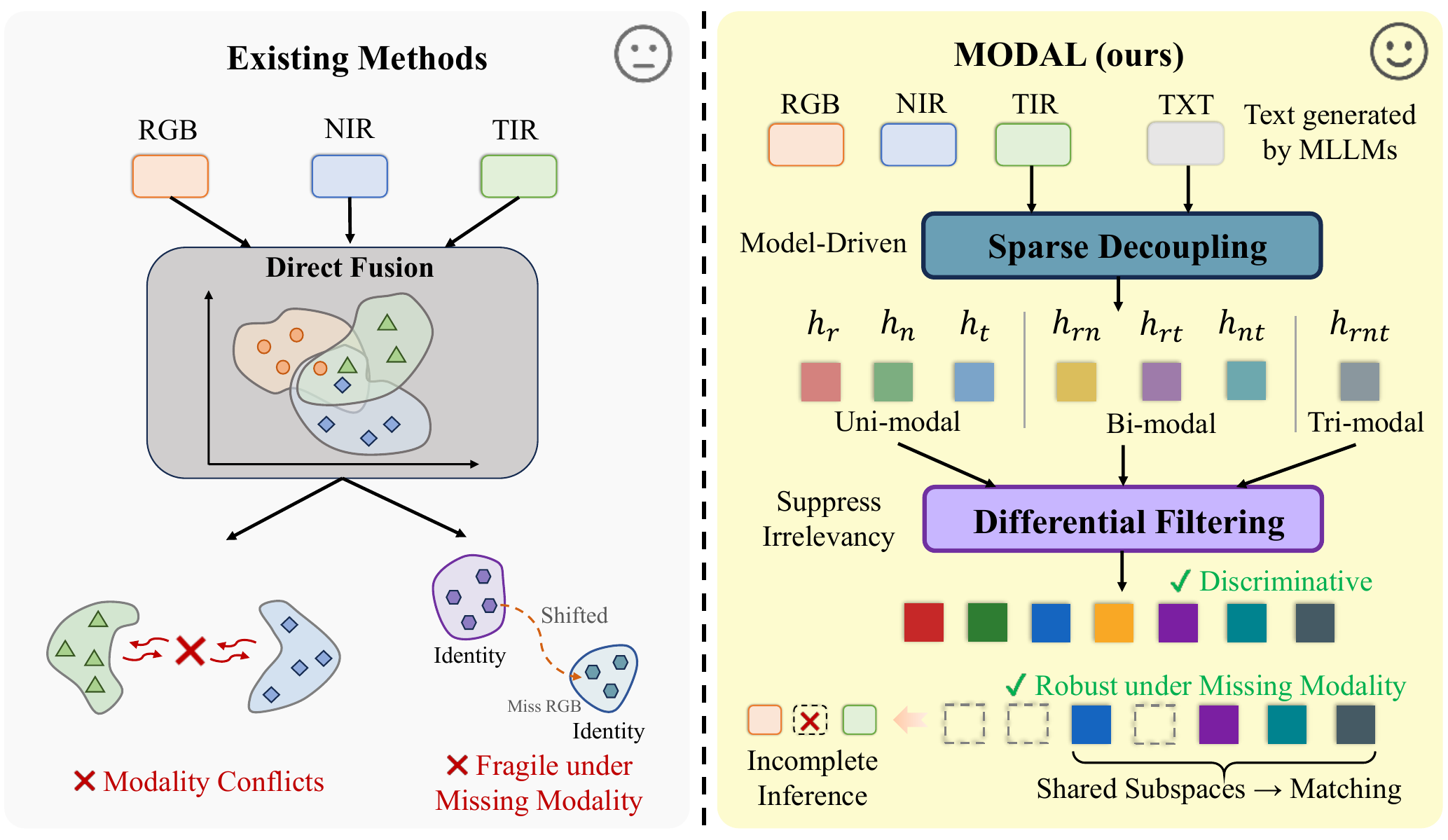}
    \caption{Motivation and overview of MODAL. Left: Existing methods directly fuse entangled multi-modal features in a unified space, which introduces inter-modal interference and leads to significant performance degradation under missing-modality scenarios. Right: In contrast, MODAL mitigates these issues through sparse feature decoupling, Text-Image Differential Filtering, and modality-aware subspace activation, effectively reducing cross-modal conflict, enhancing representation discriminability, and enabling more robust matching under incomplete-modality scenarios.}
    \label{fig. intention}
    \vspace{-6pt} 
\end{figure}

\IEEEPARstart{O}bject Re-Identification (Re-ID)~\cite{zhang2025prompt} aims to retrieve the same individual across different time, locations, sensors, and environmental conditions. As a core technology of intelligent surveillance systems, Re-ID has been widely adopted in diverse real-world applications where reliable cross-camera identity matching is critical. Traditional object Re-ID approaches~\cite{he2021transreid, miao2019pose, ye2024dynamic}, which rely on visible image for analysis, have made significant progress in addressing challenges such as variations in camera viewpoints, body poses, and occlusion. However, these methods that are solely based on visible image modality often lack sufficient effectiveness in more demanding real-world scenarios involving extreme lighting conditions, occlusions, visual camouflage, and adverse weather. Recent studies have increasingly focused on multi-modal object Re-ID~\cite{li2020multi,zheng2021robust,wang2022interact}, leveraging the complementary strengths of different modalities to improve performance in challenging scenarios. However, effectively aligning and integrating heterogeneous modal information remains challenging, as different modalities often exhibit inconsistent spatial structures, spectral distributions, and semantic granularity. More critically, most existing multi-modal models implicitly assume complete modality availability during both training and inference. In practical deployments, however, sensor failures or environmental constraints frequently render one or more modalities unavailable. Since these models directly fuse heterogeneous features into a tightly coupled joint representation, the absence of any constituent modality induces a significant distribution shift in the fused embedding, severely undermining identity similarity matching and leading to substantial performance degradation.

Consequently, the performance of multi-modal object Re-ID is fundamentally constrained by three intertwined issues: (i) the lack of principled feature disentanglement, which causes modality-specific and modality-shared information to remain entangled; (ii) direct fusion of heterogeneous modalities without structural regulation, which easily introduces cross-modal conflicts and redundant responses, thereby weakening discriminative representations; and (iii) poor robustness under modality-missing conditions, where the coupled fused representation is highly vulnerable to distribution shift once certain modalities are absent.
As illustrated in Fig.~\ref{fig. intention}(a), most existing approaches directly fuse multi-modal features in a unified embedding space without explicit structural disentanglement, making the resulting representation inherently susceptible to inter-modal interference. Moreover, such tightly coupled representations are particularly fragile under missing-modality scenarios: when one or more modalities become unavailable, the fused feature distribution shifts considerably from its complete-modality counterpart, leading to unreliable similarity estimation and significant performance degradation.

In this paper, we propose a novel Multi-modal Object Re-ID via sparse Decoupling and differentiAL filtering (MODAL) method for effectively visual and textual multi-modal features decoupling and task irrelevant feature suppression. MODAL takes multi-modal inputs, including RGB, NIR, and TIR images, complemented by textual descriptions generated by MLLMs. A Text-Image Feature Extraction module employs a pre-trained CLIP model to extract the initial text-image feature pairs. To enable transparent cross-modal modeling, we design a model-driven Multi-modal Feature Sparse Decoupling module by unfolding the algorithm of coupled sparse coding for both visual and textual multi-modal features into an interpretable deep architecture. Building upon the disentangled representations, a Text-Image Differential Filtering module leverages coarse-grained textual semantics through a differential suppression mechanism to attenuate task-irrelevant visual factors while enhancing discriminative cues. Furthermore, to address the distribution shift caused by incomplete-modality inputs, we introduce a Modality-Aware Subspace Activation (MASA). Benefiting from the principled and transparent disentanglement, MASA selectively activates only the mutually shared subspaces across available modalities, restricting inference to valid decoupled components and naturally mitigating performance degradation under modality-missing conditions.

The contributions of this paper are four-fold:
\begin{itemize}
    \item We propose \textbf{MODAL}, a principled multi-modal object Re-ID framework, establishing a unified paradigm for joint feature decoupling and fusion that generalizes to both complete- and incomplete-modality settings.

    \item We introduce the model-driven \textbf{Multi-modal Feature Sparse Decoupling (MFSD)} module, grounded incoupled sparse coding, to explicitly decompose multi-modal features.
    A \textbf{Modality-Aware Subspace Activation (MASA)} mechanism leverages this explict decoupling for robustified incomplete-modality inference.

    \item We introduce the \textbf{Text-Image Differential Filtering (TIDF)} module, which exploits coarse-grained textual description from MLLMs to suppress task-irrelevant visual responses through a multi-modal differential filtering, enhancing feature discriminability.

    \item Extensive experiments on four benchmark datasets demonstrate that MODAL consistently achieves state-of-the-art performance across standard full-modality, modality-missing, and modality-mismatched evaluation protocols, while providing structurally interpretable feature representations through its principled sparse decoupling design.
\end{itemize}

\section{Related Work}
\label{sec:relatedwork}

We briefly review three research directions closely related to our work: single-modal object re-identification, multi-modal object re-identification, and multi-modal coupled sparse coding models. 

\subsection{Single-Modal Object Re-Identification}

Early object re-identification (Re-ID) mainly relied on metric learning and hand-crafted representations~\cite{Nguyen2019KernelDM}, but such methods showed limited robustness and scalability in complex large-scale scenarios due to their insufficient capacity for modeling fine-grained identity cues~\cite{Zhu2018FastOP}. With the rise of deep learning, single-modal Re-ID has gradually shifted toward discriminative representation learning, where convolutional architectures, part-based modeling, attention mechanisms, and multi-granularity feature extraction have become the dominant paradigms for enhancing identity-related characteristics and improving feature discrimination~\cite{Wang2020ReceptiveMR,Chen2021PersonRV}. 

To further improve robustness in realistic environments, subsequent studies have extensively explored spatial misalignment, pose variation, and occlusion handling through pose-invariant representation learning, deformable part alignment, feature recovery, and context-aware reasoning~\cite{Xu2022LearningFR,Li2024OcclusionAwareTW}. More recently, Transformer-based Re-ID methods have attracted increasing attention, owing to their strong ability to model long-range dependencies and global contextual relationships. By leveraging self-attention for holistic feature interaction and flexible token aggregation, these methods provide a powerful alternative to conventional CNN-based local modeling and have shown promising performance in handling complex appearance changes, background clutter, and partial occlusion.

In parallel, unsupervised and domain-adaptive Re-ID methods have been developed to alleviate the heavy reliance on large-scale identity annotations, typically through pseudo-label learning, clustering, and cross-domain feature alignment~\cite{Li2021ClusterGuidedAC}. Beyond the conventional closed-set setting, vision-language based Re-ID have further extended the task toward more realistic long-term and open-world scenarios, where identity recognition must remain robust under substantial appearance variation and richer semantic conditions~\cite{Yang2023WinWinBC,Wei2024MultipleIP}. These advances collectively push single-modal Re-ID from conventional appearance matching toward more robust, generalizable, and semantically enriched representation learning.

\subsection{Multi-Modal Object Re-Identification}

Compared with conventional RGB-based Re-ID, cross-modal or multi-modal re-identification seeks to exploit complementary information from heterogeneous sources to improve robustness under challenging conditions such as poor illumination, occlusion, and severe appearance degradation. Existing studies have explored diverse modality combinations, including visible-infrared~\cite{Chen2022StructureAwarePT,Li2022VisibleInfraredPR,Liang2021HomogeneoustoHeterogeneousUL,Wang2026HomogeneousAH}, RGB-depth~\cite{Ren2019UniformAV}, and language-vision settings~\cite{Yan2022CLIPDrivenFT,He2023VGSGVS,Yang2023TranslationAA}. These works have substantially advanced heterogeneous Re-ID by investigating modality-specific feature extraction, cross-modal alignment, collaborative fusion, and semantic interaction strategies, with the common goal of narrowing modality gaps and learning modality-invariant identity representations.

Building upon this broader line of cross-modal Re-ID research, multi-modal object re-identification focuses on a more specific yet practically important setting, where multiple visual modalities are jointly available and need to be modeled in a unified framework. Unlike conventional cross-modal matching that mainly emphasizes alignment between two modalities, this setting further requires effective exploitation of complementarity, consistency, and interaction across multiple heterogeneous observations. Early studies primarily explored these issues by enhancing feature consistency and complementary learning among visible, near-infrared, and thermal modalities~\cite{li2020multi,zheng2021robust,wang2022interact}, establishing the foundation for subsequent multi-modal object Re-ID methods.

Recent advances have moved toward more structured and fine-grained representation learning. TOP-ReID~\cite{wang2024top} and EDITOR~\cite{zhang2024magic} improve semantic correspondence through token-level reconstruction and cross-modal token interaction, while HTT~\cite{wang2024heterogeneous} and MambaPro~\cite{wang2025mambapro} introduce hierarchical aggregation and sequence modeling to capture heterogeneous dependencies across modalities. Prompt-based methods such as PromptMA~\cite{zhang2025prompt} further explore learnable prompts to bridge modality gaps and enable adaptive feature integration. Despite their effectiveness, most existing approaches still fuse heterogeneous features within a shared embedding space, implicitly assuming that representations from different modalities are directly compatible. Such a paradigm may introduce inter-modal interference and representation conflict, thereby weakening discriminative cues and limiting generalization, especially under modality imbalance or noisy observations.

To alleviate this problem, DeMo~\cite{wang2025decoupled} explicitly decomposes multi-modal features into modality-specific and shared components with a token-based Mixture-of-Experts (MoE) design, where learnable query tokens are used to implicitly extract single modal and shared modal representations. However, its decoupling strategy remains largely heuristic, as the decomposition is driven by token learning without explicit structural priors or theoretical grounding, thus lacking principled guidance and transparency.

Moreover, current multi-modal Re-ID methods predominantly focus on visual modalities such as RGB, near-infrared, and thermal images, while largely overlooking the complementary potential of textual semantics. IDEA~\cite{wang2025idea} takes an initial step by encoding MLLM-generated text descriptions and injecting them as augmented prompts into the visual encoder. Nevertheless, such text descriptions are typically coarse-grained, summarizing only salient object attributes and background context, and are therefore insufficiently discriminative. IDEA does not explicitly consider these intrinsic limitations of the text modality, instead mainly treating it as an auxiliary semantic cue.

\begin{figure*}[t]
    \centering
    \includegraphics[width=1.95\columnwidth]{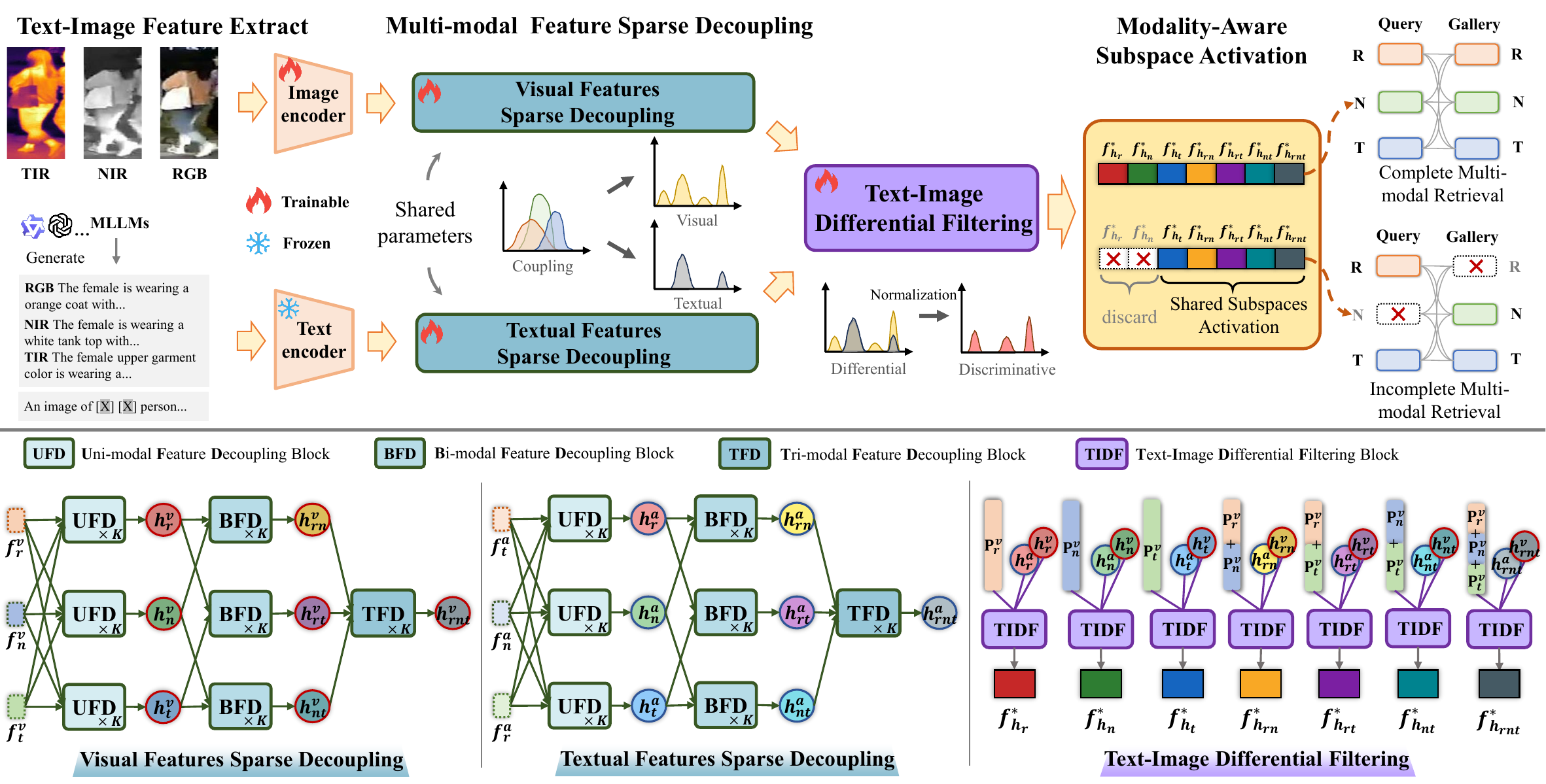}
    \caption{Overview of the proposed MODAL framework. Given multi-modal text and image inputs, their global and local features are extracted through CLIP encoders. These features are then sparsely decomposed into modality-specific and shared features via Multi-modal Feature Sparse Decoupling. Next, a Text-Image Differential Filtering module fuses the decomposed textual and visual features with local visual features to suppress task-irrelevant features and improve discriminative capability. Finally, a Mixture-of-Experts model selects and connects features to form the final representation.}
    \label{fig:overview}
    \vspace{-6pt} 
\end{figure*}

\subsection{Multi-Modal Coupled Sparse Coding Models}

The multi-modal coupled sparse coding~\cite{zhu2016coupled,gong2016coupled,wang2025decoupled,song2019coupled,mandal2016generalized,Pu2021ALB,Pu2022MixedXI,Xiong2025DFDUNDI} has demonstrated strong advantages in explicitly modeling and disentangling modality-specific and shared representations, with the applications ranging from change detection, feature selection, multi-contrast MRI reconstruction, etc.
The optimization problem is formulated with data fidelity terms with coupled dictionary and sparse codes, and sparsity constraints.

It can be solved with traditional methods, such as the Iterative Shrinkage and Thresholding Algorithm (ISTA), which are highly sensitive to hyperparameters. Gregor and LeCun~\cite{gregor2010learning} recast ISTA as a deep network by treating its dictionaries as learnable parameters. The Learnable Convolutional Sparse Coding (LCSC) method~\cite{sreter2018learned} further replaced matrix multiplications with convolution operations, which were then unfolded into deep convolutional neural networks. 
Recent works have extended the coupled sparse coding to multi-modal fusion applications. CUNet~\cite{deng2020deep} unfolds the multi-modal convolutional sparse coding model into a deep network architecture to address multi-modal image restoration and fusion tasks. InMIR-Net~\cite{deng2023interpretable} formulates multi-modal image registration as a disentangled convolutional sparse coding problem, isolating alignment-relevant features from irrelevant ones to enhance interpretability and registration accuracy. 
Furthermore, CSCFuse~\cite{zhao2023deep} employs deep convolutional sparse coding networks for diverse image fusion tasks, achieving superior performance.

Despite their success in low-level vision and image fusion tasks, coupled sparse coding models have rarely been explored in high-level recognition problems such as multi-modal object Re-ID, where discriminative representation learning under severe modality heterogeneity remains challenging.

\section{Methodology}
In this paper, we propose MODAL, a Multi-modal Object Re-ID framework via sparse Decoupling and differential filtering. As illustrated in Fig.~\ref{fig:overview}, MODAL comprises four key components: (i) a Text-Image Feature Extraction module that obtains aligned visual and textual representations from pre-trained CLIP encoders; (ii) a Multi-modal Feature Sparse Decoupling module that decomposes multi-modal visual and textual features into uni-modal specific, bi-modal shared, and tri-modal shared components via model-driven deep unrolling; (iii) a Text-Image Differential Filtering module that leverages decoupled textual semantics to suppress task-irrelevant visual responses; and (iiii) a Modality-Aware Subspace Activation mechanism that selectively activates shared subspaces for robust matching under both full- and incomplete-modality scenarios.

\subsection{Text-Image Feature Extraction}
A Text-Image Feature Extraction module is introduced to extract semantically aligned visual and textual representations from multi-modal images and their corresponding MLLM-generated descriptions.
We denote the image input as $\mathbf{X}^v_m $ with the superscript \textit{v} referring to visual content, and $m \in \{ r, n, t \}$ corresponds to RGB, NIR, and TIR images, respectively. Following~\cite{hu2024empowering,wang2025idea}, each image input $\mathbf{X}^v_m$ is processed by MLLMs to generate textual summaries. Learnable prompts prefix is prepended to each summary, and the implicit prefix is concatenated with the explicit textual content, forming a comprehensive text annotation $\mathbf{X}^a_m$ for each modality with superscript \textit{a} for text annotation.
To establish semantically aligned representations, we extract visual and textual features from pre-trained CLIP encoders.
We denote the visual feature and textual feature as $\mathbf{F}_m^v=[\bm{f}^v_{m},\mathbf{P}^v_{m}]$ and $\mathbf{F}^a_{m}=[\bm{f}^a_{m},\mathbf{P}^a_{m}]$, respectively, where $\bm{f}$ denotes the class token and $\mathbf{P}$ represents the patch tokens.

\subsection{Multi-modal Feature Sparse Decoupling}
A Multi-modal Feature Sparse Decoupling (MFSD) module grounded in coupled sparse coding theory is proposed to achieve principled and transparent disentanglement of multi-modal representations through a model-driven deep unrolling architecture.

\begin{figure}[t]
    \centering
    \includegraphics[width=0.95\columnwidth]{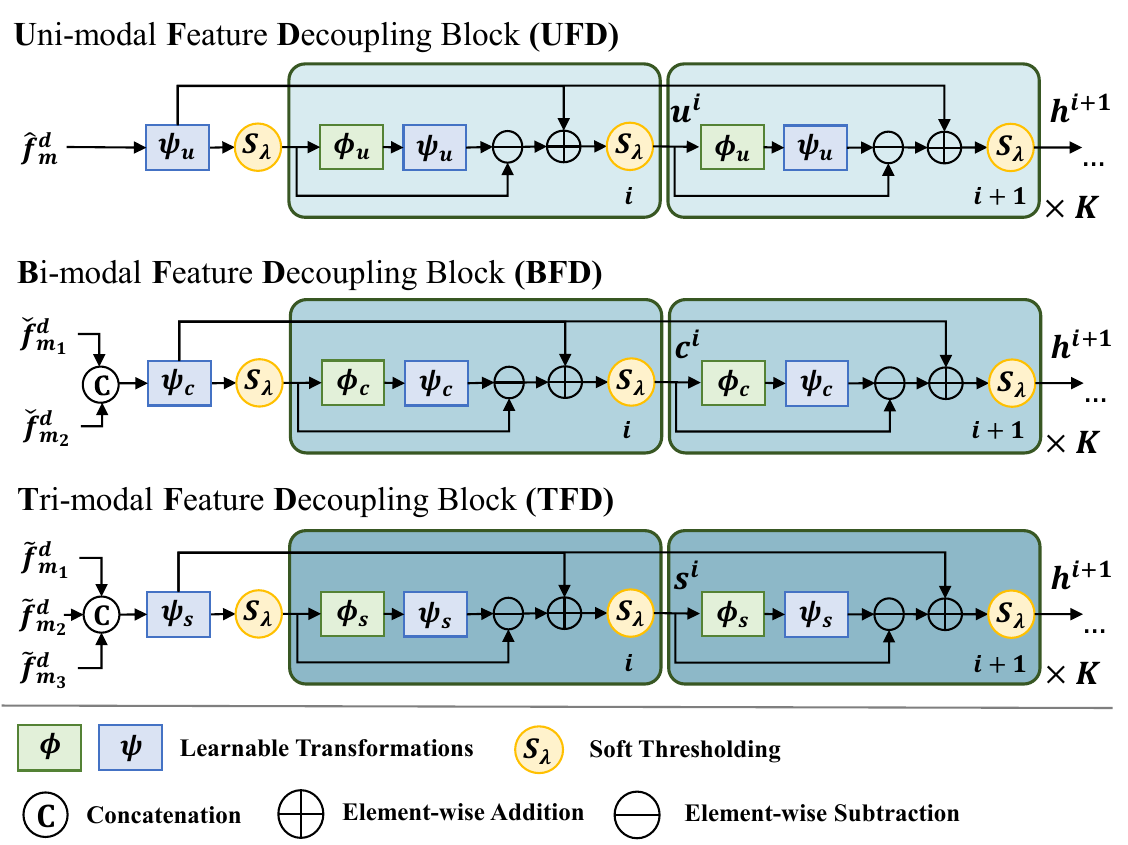}
    \caption{Architecture of the Uni-modal (UFD), Bi-modal (BFD), and Tri-modal (TFD) feature decoupling blocks. All three blocks utilize a unified iterative structure. Each block refines its corresponding sparse codes through $K$ repeated stages of learnable transformations and soft thresholding.}
    \label{fig:MFSD}
    \vspace{-6pt} 
\end{figure}

\noindent\textbf{Overall MFSD Architecture:}
As illustrated in Fig.~\ref{fig:overview}, MFSD consists of three sets of Uni-modal Feature Decoupling (UFD) blocks, three sets of Bi-modal Feature Decoupling (BFD) blocks and one set of Tri-modal Feature Decoupling (TFD) blocks to sequentially optimize the uni-modal sparse features $\{\bm{h}_r, \bm{h}_n, \bm{h}_t\}$, bi-modal sparse features $\{\bm{h}_{rn}, \bm{h}_{rt}, \bm{h}_{nt}\}$, and the tri-modal sparse feature ${h}_{rnt}$.
The parameters of the MFSD module is shared for both image and text modalities to ensure consistent semantic alignment across modalities.
Through this progressive and iterative decoupling process, MFSD produces disentangled features $\mathcal{Z}$ encompassing both image and text modalities, which serves as the foundation for the subsequent Text-Image Differential Filtering.
The detailed structures of 
UFD, BFD, and TFD blocks are given in Fig.~\ref{fig:MFSD}.

\noindent\textbf{Problem Formulation:} 
We assume that heterogeneous multi-modal features can be linearly represented from modality-specific and shared sparse bases.
We denote the class tokens as $\bm{f}_m^d$ , where $m \in \{ r,n,t \}$ indicates the modality (RGB, NIR, or TIR) and $d \in \{ v, a \}$ denotes the feature domain (visual or textual).
Each class token is represented as a linear combination of three components: the uni-modal specific feature $\{\bm{h}_{m} \}$, the bi-modal shared features $\{\bm{h}_{m_1 m_2}\}$ with $m_i \in \{ r, n, t \}$, and the tri-modal shared feature ${h}_{rnt}$:
\begin{equation}
    \begin{cases}
    \bm{f}^d_{r} = \mathbf{R}_r \bm{h}_{r} + \mathbf{R}_{rn} \bm{h}_{rn} + \mathbf{R}_{rt} \bm{h}_{rt} + \mathbf{R}_s \bm{h}_{rnt}, \\
    \bm{f}^d_{n} = \mathbf{N}_{n} \bm{h}_{n} + \mathbf{N}_{rn} \bm{h}_{rn} + \mathbf{N}_{nt} \bm{h}_{nt} + \mathbf{N}_s \bm{h}_{rnt}, \\
    \bm{f}^d_{t} = \mathbf{T}_t \bm{h}_{t} + \mathbf{T}_{rt} \bm{h}_{rt} + \mathbf{T}_{nt} \bm{h}_{nt} + \mathbf{T}_s \bm{h}_{rnt},
\end{cases}
\label{eq1}
\end{equation}
where $\mathbf{R, N, T}$ denotes the sparse dictionaries for RGB, NIR, and TIR, respectively, the subscripts $r$, $n$ and $t$ correspond to the RGB, NIR, and TIR modalities.

To optimize the set of seven features $\mathcal{Z} = \left\{ \bm{h}_{r}, \bm{h}_{n}, \bm{h}_{t}, \bm{h}_{rn}, \bm{h}_{rt}, \bm{h}_{nt}, \bm{h}_{rnt} \right\}$, we formulate a sparse optimization problem with the above seven optimization variables:
\begin{equation}
\small
\begin{aligned}
\arg\mathop{\min}\limits_{\mathcal{Z}} \frac{1}{2} \Big(
& \left\|\bm{f}^d_{r} - \mathbf{R}_r \bm{h}_{r} - \mathbf{R}_{rn} \bm{h}_{rn} - \mathbf{R}_{rt} \bm{h}_{rt} - \mathbf{R}_s \bm{h}_{rnt}\right\|_2^2 \\
+\, & \left\|\bm{f}^d_{n} - \mathbf{N}_n \bm{h}_{n} - \mathbf{N}_{rn} \bm{h}_{rn} - \mathbf{N}_{nt} \bm{h}_{nt} - \mathbf{N}_s \bm{h}_{rnt}\right\|_2^2 \\
+\, & \left\|\bm{f}^d_{t} - \mathbf{T}_t \bm{h}_{t} - \mathbf{T}_{rt} \bm{h}_{rt} - \mathbf{T}_{nt} \bm{h}_{nt} - \mathbf{T}_s \bm{h}_{rnt}\right\|_2^2 \Big) \\
+\, & \lambda \sum_{\bm{z} \in \mathcal{Z}} \|\bm{z}\|_1
\end{aligned}
\label{eq2}
\end{equation}
where $\lambda$ is a regularization parameter to balance the data fidelity terms and the sparse prior term.

The objective is to decouple the multi-modal features into the set of decoupled sparse features $\mathcal{Z} = \left\{ \bm{h}_{r}, \bm{h}_{n}, \bm{h}_{t}, \bm{h}_{rn}, \bm{h}_{rt}, \bm{h}_{nt}, \bm{h}_{rnt} \right\}$ therefore preventing the redundant feature stacking and exploiting the complementary features. 
Such multi-variable sparse coding problem can be solved using an alternating update strategy, which iteratively updates one variable while keeping the others fixed. The details of uni-modal feature decoupling block are introduced as follows.

\noindent\textbf{Uni-modal Feature Decoupling (UFD) Block: }
There are three UFD blocks for updating the uni-modal sparse features $\{\bm{h}_r, \bm{h}_n, \bm{h}_t\}$. We take the UFD block design of $\bm{h}_r$ as an example for illustration.
Let $\hat{\bm{f}}^d_{r}=\bm{f}^d_{r}-(\mathbf{R}_{rn}\bm{h}_{rn}+\mathbf{R}_{rt}\bm{h}_{rt}+\mathbf{R}_s\bm{h}_{rnt})$, the optimization problem for the uni-modal feature $\bm{h}_r$ can be formulated as:
\begin{align}
    \arg\mathop{\min}\limits_{\bm{h}_{r}} \frac{1}{2}\left\|\bm{\bm{\hat{f}}}^d_{r}-\mathbf{R}_r \bm{h}_{r}\right\|_2^2+\lambda \|\bm{h}_{r}\|_1.
    \label{eq2}
\end{align}

This optimization problem can be solved in an iterative manner. The iterative algorithm for solving Eqn. (\ref{eq2}) has been unrolled into $K$ UFD blocks and
the $i$-th UFD block can be expressed as:
\begin{equation}
    \bm{h}_{r}^{i+1}=\mathcal{S}_\lambda\left(\bm{h}_{r}^i-\mathbf{\psi}_u( \mathbf{\phi}_u (\bm{h}_{r}^i))+ \mathbf{\psi}_u(\hat{\bm{f}}^d_{r})\right),
\label{eq3}
\end{equation}
where $\mathcal{S}_\lambda(\cdot)$ denotes the soft-thresholding operator with learnable threshold $\lambda$, and $\phi_u(\cdot),\psi_u(\cdot)$ are learnable transformations implemented using a bottleneck structure composed of linear layers and dropout-based nonlinear activations. This design achieves balance between expressiveness and efficiency, yielding the best performance among several alternative implementations.

\noindent\textbf{Bi-modal Feature Decoupling (BFD) Block: }
There are three set of BFD blocks within MFSD module for the bi-modal shared sparse features $\{ \bm{h}_{rn}, \bm{h}_{rt}, \bm{h}_{nt}\}$.
Take optimizing the variable $\bm{h}_{rn}$ as an example, the optimization problem for the bi-modal feature $\bm{h}_{rn}$ can be reformulated as:
\begin{equation}
\begin{split}
    \arg\mathop{\min}\limits_{\bm{h}_{rn}} & \frac{1}{2}\left\|\bm{\check{f}}^d_{r}-\mathbf{R}_{rn}\bm{h}_{rn}\right\|_2^2 + \frac{1}{2}\left\|\bm{\check{f}}^d_{n}-\mathbf{N}_{rn}\bm{h}_{rn}\right\|_2^2\\
    &+\lambda \|\bm{h}_{rn}\|_1,
    \label{eq Sup 1}
\end{split}
\end{equation}
where $\check{\bm{f}}^d_{r}=\bm{f}^d_{r}-(\mathbf{R}_r\bm{h}_r+\mathbf{R}_{rt}\bm{h}_{rt}+\mathbf{R}_s\bm{h}_{rnt})$, and $\check{\bm{f}}^d_{n}=\bm{f}^d_{n}-(\mathbf{N}_{n}\bm{h}_n+\mathbf{N}_{nt}\bm{h}_{nt}+\mathbf{N}_s\bm{h}_{rnt})$, and $\lambda$ is a regularization parameter.

By combining the first two terms in Eqn. (\ref{eq Sup 1}), the optimization problem can be reformulated as follows:
\begin{equation}
\begin{split}
    &\arg\mathop{\min}\limits_{\bm{h}_{rn}} \frac{1}{2}\left\|[\bm{\check{f}}^d_{r},\bm{\check{f}}^d_{n}]-\mathbf{E}_{rn}\bm{h}_{rn}\right\|_2^2 +\lambda \|\bm{h}_{rn}\|_1,
    \label{eq Sup 2}
\end{split}
\end{equation}
where $\mathbf{E}_{rn}$ is the concatenation of $\mathbf{R}_{rn}$ and $\mathbf{N}_{rn}$.

Therefore, we formulate a standard sparse optimization problem, solvable via a closed-form solution based on LISTA algorithm~\cite{gregor2010learning}:
\begin{equation}
    \bm{h}_{rn}^{i+1}=\mathcal{S}_\lambda\left(\bm{h}_{rn}^i-\mathbf{\psi}_c( \mathbf{\phi}_c (\bm{h}_{rn}^i)) + \mathbf{\psi}_c([\bm{\check{f}}^d_{r},\bm{\check{f}}^d_{n}])\right),
\label{eq Sup 3}
\end{equation}
where $\mathcal{S}_\lambda(\cdot)$ represents a soft-thresholding operator, $\phi_c(\cdot),\psi_c(\cdot)$ denotes learnable transformations, and we concretely implement them using a bottleneck structure composed of linear layers and dropout-based nonlinear activations.

\noindent\textbf{Tri-modal Feature Decoupling (TFD) Block: }
For the tri-modal feature decoupling, the optimization problem for the tri-modal feature ${h}_{rnt}$ can be reformulated as:
\begin{equation}
\begin{split}
    \arg\mathop{\min}\limits_{\bm{h}_{rnt}} &\frac{1}{2}\left\|\tilde{\bm{f}}^d_{r}-\mathbf{{R}}_s\bm{h}_{rnt}\right\|_2^2 + \frac{1}{2}\left\|\tilde{\bm{f}}^d_{n}-\mathbf{{N}}_s\bm{h}_{rnt}\right\|_2^2 \\
    &+ \frac{1}{2}\left\|\tilde{\bm{f}}^d_{t}-\mathbf{{T}}_s\bm{h}_{rnt}\right\|_2^2 +\lambda \|\bm{h}_{rnt}\|_1,
    \label{eq Sup 4}
\end{split}
\end{equation}
where $\tilde{\bm{f}}^d_{r}=\bm{f}^d_{r}-(\mathbf{R}_r\bm{h}_r+\mathbf{R}_{rn}\bm{h}_{rn}+\mathbf{R}_{rt}\bm{h}_{rt})$, $\tilde{\bm{f}}^d_{n}=\bm{f}^d_{n}-(\mathbf{N}_{n}\bm{h}_n+\mathbf{N}_{rn}\bm{h}_{rn}+\mathbf{N}_{nt}\bm{h}_{nt})$, and $\tilde{\bm{f}}^d_{t}=\bm{f}^d_{t}-(\mathbf{T}_t\bm{h}_t+\mathbf{T}_{rt}\bm{h}_{rt}+\mathbf{T}_{nt}\bm{h}_{nt})$, and $\lambda$ is a regularization parameter.

By combining the first three terms in Eqn. (\ref{eq Sup 4}), the optimization problem can be reformulated as follows:
\begin{equation}
\begin{split}
    &\arg\mathop{\min}\limits_{{h}_{rnt}} \frac{1}{2}\left\|[\bm{\tilde{f}}^d_{r},\bm{\tilde{f}}^d_{n},\bm{\tilde{f}}^d_{t}]-\mathbf{G}_{s}\bm{h}_{rnt}\right\|_2^2 +\lambda \|\bm{h}_{rnt}\|_1,
    \label{eq Sup 5}
\end{split}
\end{equation}
where $\mathbf{G}_{s}$ is the concatenation of $\mathbf{R}_{s}$, $\mathbf{N}_{s}$ and $\mathbf{T}_{s}$. 

Thus, we can express the closed-form solution of the tri-modal shared feature as:
\begin{equation}
    {h}_{rnt}^{i+1}=\mathcal{S}_\lambda\left(\bm{h}_{rnt}^i-\mathbf{\psi}_s( \mathbf{\phi}_s( \bm{h}_{rnt}^i))+ \mathbf{\psi}_s([\bm{\tilde{f}}^d_{r},\bm{\tilde{f}}^d_{n},\bm{\tilde{f}}^d_{t}])\right),
\label{eq5}
\end{equation}
where $\mathcal{S}_\lambda(\cdot)$ represents a soft-thresholding operator, $\phi_s(\cdot),\psi_s(\cdot)$ denotes learnable transformations, and we concretely implement them using a bottleneck structure composed of linear layers and dropout-based nonlinear activations.

\subsection{Text-Image Differential Filtering}
We observe that the text descriptions generated by Multi-modal Large Language Models are coarse-grained and are not sufficiently discriminative, as reflected in Table \ref{tab. ablation text and image usage},
yet they may offer semantic cues that can guide the suppression of task irrelevant or disruptive features in visual features.

We propose a Text-Image Differential Filtering (TIDF) module to effectively fuse visual-textual features building on this insight. 
The proposed TIDF module draws inspiration from the differential transformer~\cite{Ye2024DifferentialT}, which filters irrelevant signals by mapping a single feature into two query projections and computing their differential attention to suppress noisy channel responses. In contrast, our TIDF module maps the decoupled visual and textual features into separate query embeddings, exploiting the coarse-grained textual semantics to identify and suppress task-irrelevant or disruptive responses in the visual representations, thereby enhancing discriminative patterns.

\begin{figure}[t]
    \centering
    \includegraphics[width=0.75\columnwidth]{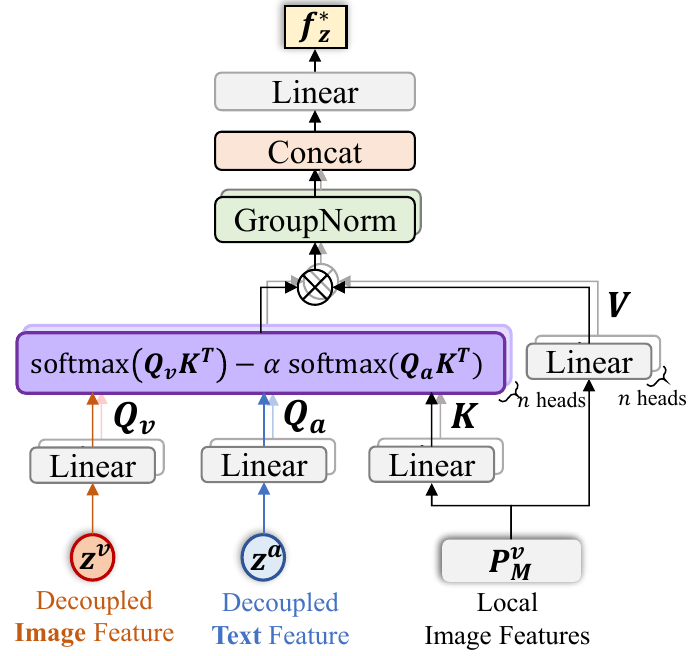}
    \caption{Architecture of the Text-Image Differential Filtering block.}
    \label{Fig.TIDF}
\end{figure}

Specifically, given the the local image features $\mathbf{P}_{M}^v$ extracted from CLIP ($M$ indicates the modality combination associated with the current decoupled feature), the TIDF block performs a differential filtering between its corresponding text-image feature pair $(\bm{z}^v, \bm{z}^a)$ and $\mathbf{P}_{M}^v$. As illustrated in Fig.~\ref{Fig.TIDF}, each TIDF block first projects the decoupled textual and visual features into query embeddings $\mathbf{Q}_v$ and $\mathbf{Q}_a$ through a linear transformation, while the corresponding image patch tokens $\mathbf{P}_{M}^v$ are mapped into key–value pairs $(\mathbf{K}, \mathbf{V})$. The response $\mathbf{Q}_v\mathbf{K}^\top$ captures the visual activations, whereas $\mathbf{Q}_a\mathbf{K}^\top$ encodes task-irrelevant or disruptive cues guided by textual semantics. Their differential filtering enhances the discriminative capability of the feature representations. The resulting attention map is then multiplied with $\mathbf{V}$, followed by group normalization, concatenation across heads, and a final linear projection to produce the fused representation $\bm{f}^*_{\bm{z}}$ corresponding to the feature subset $\bm{z} \in \mathcal{Z}$. Formally, the differential filtering in each block can be expressed as:
\begin{equation}
\small
\begin{cases}
    \bm{q}_v = \bm{z}^v \mathbf{W}^{Q_v}, \quad \bm{q}_a = \bm{z}^a \mathbf{W}^{Q_a},\\[4pt]
    \mathbf{K} =\mathbf{P}_{M}^v \mathbf{W}^{K}, \quad \mathbf{V} = \mathbf{P}_{M}^v\mathbf{W}^{V},\\[4pt]
    \bm{f}^*_{\bm{z}} = \left(\mathrm{softmax}\!\left(\frac{\bm{q}_v\mathbf{K}^T}{\sqrt{d}}\right) - \alpha\, \mathrm{softmax}\!\left(\frac{\bm{q}_a\mathbf{K}^T}{\sqrt{d}}\right)\right)\mathbf{V},
\end{cases}
\label{eq6}
\end{equation}
where $\mathbf{W^{Q_v}}$, $\mathbf{W^{Q_a}}$, $\mathbf{W^{K}}$ and, $\mathbf{W^{V}}$ denote the linear transforms. 
Moreover, $\alpha$ is a learnable parameter, and is re-parameterized as:
\begin{equation}
    \alpha=\exp(\alpha_{q1}\cdot\alpha_{k1})-\exp(\alpha_{q2}\cdot\alpha_{k2})+\alpha_{\mathrm{init}},
\end{equation}
where $\alpha_{q1}, \alpha_{k1}, \alpha_{q2}, \alpha_{k2}$ are learnable vectors, and $\alpha_{\mathrm{init}}$ is a constant.

Finally, all of the fused representations $\bm{f}^*_{\bm{z}}$ obtained from the corresponding TIDF blocks are each fed into an independent Mixture-of-Experts (MoE) module equipped with adaptive learnable gating. The outputs of all MoE modules are then concatenated to form the fused representation $\bm{F}^*$. The ultimate object feature is obtained as ${\bm{\hat{F}}}^* = [\bm{f}_m^d, \bm{F}^*]$. As shown in Fig.~\ref{fig:overview}, there are seven parallel TIDF blocks, each corresponding to one of the seven decoupled sparse feature sets $\bm{z} \in \left\{ \bm{h}_{r}, \bm{h}_{n}, \bm{h}_{t}, \bm{h}_{rn}, \bm{h}_{rt}, \bm{h}_{nt}, \bm{h}_{rnt} \right\}$ obtained by the MFSD module. Each TIDF block performs a differential filtering between its corresponding decoupled text-image feature pair and the local image patch tokens, which serve as keys and values to enable fine-grained spatial feature extraction.

\subsection{Modality-Aware Subspace Activation}
\label{sec:masa}
We further propose a Modality-Aware Subspace Activation~(MASA) mechanism to bridge the gap between full-modality and incomplete-modality inference. While the MFSD and TIDF modules described above yield a set of structurally disentangled and semantically refined feature components, real-world deployments often suffer from sensor failures, occlusions, or asynchronous acquisition that render one or more modalities unavailable. This causes the modality configurations of query and gallery samples to differ, undermining the reliability of standard matching procedures. MASA addresses this challenge by leveraging the explicit structure of the disentangled representation, selectively activating only the valid subspaces shared between any given query-gallery pair. This unifies full-modality and incomplete-modality inference within a single, coherent framework without requiring auxiliary reconstruction or modality hallucination.

The central observation is that, owing to the principled decomposition enforced by MFSD, each component in $\mathcal{Z} = \{\bm{h}_r, \bm{h}_n, \bm{h}_t, \bm{h}_{rn}, \bm{h}_{rt}, \bm{h}_{nt}, \bm{h}_{rnt}\}$ is associated with a clearly defined subset of source modalities.
We formalize this association through the notion of \emph{modality support}:
\begin{equation}
	\mathcal{S}(\bm{h}_m) = \{m\},\quad
	\mathcal{S}(\bm{h}_{ij}) = \{i, j\},\quad
	\mathcal{S}(\bm{h}_{rnt}) = \{r, n, t\},
	\label{eq:support}
\end{equation}
where $m \in \{r, n, t\}$ and $\{i,j\} \subset \{r,n,t\}$ with $i \neq j$.
The modality support specifies the minimal set of modalities that must be present for a component to carry meaningful information.

Given a query--gallery pair $(q, g)$ with available modality sets $\mathcal{M}_q \subseteq \{r, n, t\}$ and $\mathcal{M}_g \subseteq \{r, n, t\}$, a component $z$ is deemed \emph{valid} if and only if its modality support is jointly covered by both samples:
\begin{equation}
	\mathcal{S}(\bm{z}) \subseteq \mathcal{M}_q \cap \mathcal{M}_g.
	\label{eq:validity}
\end{equation}
The activated component set is accordingly defined as:
\begin{equation}
	\mathcal{Z}_{qg} = \bigl\{\, \bm{z} \in \mathcal{Z} \;\big|\; \mathcal{S}(\bm{z}) \subseteq \mathcal{M}_q \cap \mathcal{M}_g \,\bigr\}.
	\label{eq:activated}
\end{equation}
When $\mathcal{M}_q = \mathcal{M}_g = \{r,n,t\}$, all seven components are activated and the full representation participates in matching.
When one or more modalities are absent, components whose modality support exceeds the available set are automatically deactivated, and matching proceeds exclusively over mutually valid subspaces.

Let $\hat{\bm{F}}^{*} = [\bm{f}^*_{u_r}, \bm{f}^*_{u_n}, \bm{f}^*_{u_t}, \bm{f}^*_{c_{rn}}, \bm{f}^*_{c_{rt}}, \bm{f}^*_{c_{nt}}, \bm{f}^*_{s}]$ denote the composite representation after TIDF, where each $\bm{f}^*_z$ is the fused feature corresponding to component $z$.
MASA retains only the valid components via block-wise selection:
\begin{equation}
	\tilde{\bm{F}}_{qg} = \operatorname*{Concat}_{\bm{z} \,\in\, \mathcal{Z}_{qg}} \bm{f}^*_{\bm{z}}\,,
	\label{eq:selection}
\end{equation}
where $\operatorname{Concat}$ denotes concatenation over the activated components.
The similarity between $q$ and $g$ is then computed by aggregating the component-wise inner products:
\begin{equation}
	\mathrm{Sim}(q, g) = \sum_{\bm{z} \,\in\, \mathcal{Z}_{qg}} \bigl\langle \bm{f}^{*(q)}_{\bm{z}},\; \bm{f}^{*(g)}_{\bm{z}} \bigr\rangle.
	\label{eq:similarity}
\end{equation}

Three properties of this formulation are worth highlighting.
First, because MFSD explicitly enforces sparse structural disentanglement, each component resides in a semantically distinct subspace; deactivating a component therefore does not introduce interference into the remaining ones.
Second, unlike reconstruction-based strategies that hallucinate missing-modality features, MASA operates directly on the structurally consistent components without any auxiliary generation or adaptation, thereby avoiding error propagation from imprecise imputation.
Third, in practical retrieval the gallery features $\{\bm{f}^{*(g)}_{\bm{z}}\}_{\bm{z} \in \mathcal{Z}}$ are pre-computed offline for all seven components; at query time, $\mathcal{Z}_{qg}$ is determined on-the-fly and only the corresponding similarities are accumulated, incurring negligible additional cost.

\begin{figure}[t]
    \centering
    \includegraphics[width=0.99\columnwidth]{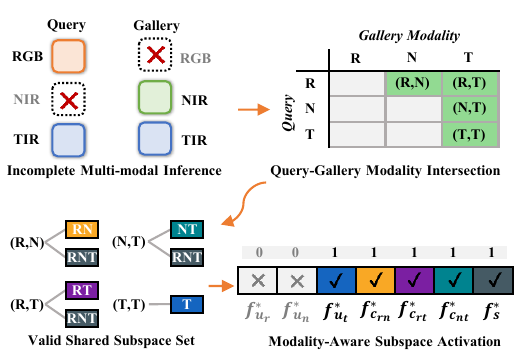}
    \caption{Schematic illustration of the Modality-Aware Subspace Activation (MASA) for incomplete modality inference..}
    \label{Fig.MASA}
\end{figure}

\subsection{Loss Functions}
We optimize the network using a combination of identity classification loss and metric learning loss. The overall loss is defined as:
\begin{equation}
\mathcal{L}=\mathcal{L}_{tri}(\bm{\tilde{F}})
+\eta\mathcal{L}_{id}(\bm{\tilde{F}}),
\label{eq: loss}
\end{equation}
where $\bm{\tilde{F}}$ denotes the modality-aware feature embedding after MASA selection as defined in Eq.~\eqref{eq:selection}, $\mathcal{L}_{id}$ is the cross-entropy loss with label smoothing, $\mathcal{L}_{tri}$ is the triplet loss~\cite{hermans2017defense}, and $\eta$ is a balancing coefficient.

\section{Experiments}

We conduct comprehensive experiments on four tri-modal object Re-ID benchmarks spanning both pedestrian and vehicle domains, evaluating MODAL under standard full-modality, modality-missing, and modality-mismatched protocols to validate its effectiveness, and robustness.

\subsection{Datasets and Evaluation Protocols} 
\subsubsection{Datasets} We evaluate the proposed method on four representative tri-modal object Re-ID benchmarks, including one pedestrian dataset and three vehicle datasets.

\textbf{RGBNT201}~\cite{zheng2021robust} is a tri-modal person Re-ID dataset containing 4,787 aligned RGB--NIR--TIR image triples from 201 identities captured across multiple cameras, serving as a standard benchmark for multi-modal pedestrian re-identification.

\textbf{RGBNT100}~\cite{li2020multi} is a large-scale tri-modal vehicle Re-ID dataset comprising 17,250 RGB--NIR--TIR image triples collected under diverse illumination conditions, providing extensive cross-spectral variations for vehicle identity matching.

\textbf{MSVR310}~\cite{zheng2023cross} contains 2,087 aligned tri-modal image triples from 310 vehicle identities captured across different environments, serving as a compact yet challenging benchmark for cross-domain vehicle Re-ID.

\textbf{WMVEID863}~\cite{zheng2025flare} consists of 863 vehicle identities with significant illumination variations, enabling evaluation of cross-spectral generalization capability under diverse lighting conditions.

For each image, text descriptions are generated by Multi-modal Large Language Models (MLLMs). Specifically, the text annotations of RGBNT201, RGBNT100, and MSVR310 are directly adopted from the resources released by IDEA~\cite{wang2025idea}, while those for WMVEID863 are generated following the same MLLM-based procedure described in IDEA.

\subsubsection{Evaluation Protocols.} Following standard practices in the ReID community, we apply Rank-$K$ ($K = 1, 5, 10$) matching accuracy and mean Average Precision (mAP) to evaluate
the model’s accuracy and generalization performance. All evaluation protocols are kept consistent with those used in prior comparison works to ensure fair and reliable results.

Besides the standard evaluation protocols, we further consider two extended evaluation settings, namely modality-missing and modality-mismatched scenarios, to assess the robustness of multi-modal ReID models. Detailed descriptions of these protocols are provided in Section~\ref{sec:Miss modal}.

\subsection{Implementation Details}
\noindent The MODAL is implemented in PyTorch. Input images are resized to 256×128 for RGBNT201 and 128×256 for RGBNT100, MSVR310 and WMVEID863. Following common practice, we employ random horizontal flipping, cropping, and erasing for data augmentation. We adopt the Adam optimizer with a base learning rate of $1e^{-5}$, and follow by a cosine decay to $1e^{-6}$. The mini-batch configuration varies across datasets: for RGBNT201, MSVR310 and WMVEID863, we use a batch size of 64, sampling 8 images per identity, whereas for RGBNT100, we increase the batch size to 128 with 16 images per identity. The model is trained for 50 epochs in total. The number of experts is set to 7 according to the number of TIDF blocks; the parameter $\eta$ in E.q~\ref{eq: loss} is set as 0.25, empirically.
All experiments were conducted on a computer equipped with an NVIDIA 4090 GPU.

\begin{table}[t]
    \centering
    \caption{Comparison of different methods on pedestrian dataset RGBNT201.
    ($\dagger$ denotes the methods that additionally utilize text annotations from {IDEA}~\cite{wang2025idea}.)
    }
    \label{tab.SOTA pedestrian}
    \resizebox{0.46\textwidth}{!}{
    \begin{tabular}{llccc}
        \toprule
        \multirow{2}{*}{\textbf{Methods}} & \multirow{2}{*}{\textbf{Venue}} & \multicolumn{3}{c}{\textbf{RGBNT201}} \\
        \cmidrule(lr){3-5} 
        & & \textbf{mAP} & \textbf{Rank-1} & \textbf{Rank-5}\\
        \midrule
        HAMNet \cite{li2020multi} & AAAI-20 & 27.7 & 26.3 & 41.5\\
        PFNet \cite{zheng2021robust} & AAAI-21 & 38.5 & 38.9 & 52.0\\
        IEEE \cite{wang2022interact} & AAAI-22 & 46.4 & 47.1 & 58.5\\
        LRMM \cite{wu2025lrmm} & ESWA-25 & 52.3 & 53.4 & 64.6\\
        TIENet \cite{yang2025tienet} & TNNLS-25 & 54.4 & 54.4 & 66.3\\
        EDITOR \cite{zhang2024magic} & CVPR-24 & 66.5 & 68.3 & 81.1\\
        WTSF-ReID \cite{yu2025wtsf} & ESWA-25 & 67.9 & 72.2 & 83.4\\
        HTT \cite{wang2024heterogeneous} & AAAI-24 & 71.1 & 73.4 & 83.1\\
        TOP-ReID \cite{wang2024top} & AAAI-24 & 72.3 & 76.6 & 84.7\\
        ICPL-ReID \cite{li2026icpl} & TMM-25 & 75.1 & 77.4 & 84.2\\
        PromptMA \cite{zhang2025prompt} & TIP-25 & 78.4 & 80.9 & 87.0\\
        MambaPro \cite{wang2025mambapro} & AAAI-25 & 78.9 & 83.4 & 89.8\\
        DeMo \cite{wang2025decoupled} & AAAI-25 & 79.0 & 82.3 & 88.8\\
        $\text{IDEA}^{\dagger}$ \cite{wang2025idea} & CVPR-25 & 80.2 & 82.1 & 90.0\\
        Signal \cite{liu2026signal} & AAAI-26 & \cellcolor{secondblue}80.3 & \cellcolor{secondblue}85.2 & \cellcolor{bestblue}\textbf{91.4}\\
        $\text{MODAL}^{\dagger}$ & Ours & \cellcolor{bestblue}\textbf{83.8} & \cellcolor{bestblue}\textbf{86.1} & \cellcolor{secondblue}90.3\\
        \noalign{\hrule height 1pt}
    \end{tabular}
    }
\end{table}

\begin{table*}[t]
    \centering
    \caption{Comparison of different methods on three vehicle datasets: RGBNT100, MSVR310, and WMVEID863. ($\dagger$ denotes the methods that additionally utilize text annotations from {IDEA}~\cite{wang2025idea}.)
    }
    \label{tab.SOTA vehicle}
    \resizebox{0.92\textwidth}{!}{
    \begin{tabular}{llcccccccccc}
        \toprule
        \multirow{2}{*}{\textbf{Methods}} & \multirow{2}{*}{\textbf{Venue}} & \multicolumn{3}{c}{\textbf{RGBNT100}} & \multicolumn{3}{c}{\textbf{MSVR310}} & \multicolumn{4}{c}{\textbf{WMVEID863}} \\
        \cmidrule(lr){3-5} \cmidrule(lr){6-8} \cmidrule(lr){9-12}
        & & \textbf{mAP} & \textbf{Rank-1} & \textbf{Rank-5} & \textbf{mAP} & \textbf{Rank-1} & \textbf{Rank-5} & \textbf{mAP} & \textbf{Rank-1} & \textbf{Rank-5} & \textbf{Rank-10}\\
        \midrule
        IEEE \cite{wang2022interact} & AAAI-22 & 61.3 & 87.8 & 90.2 & 21.0 & 41.0 & 57.7 & 45.9 & 48.6 & 64.3 & 67.9\\
        PFNet \cite{zheng2021robust} & AAAI-21 & 68.1 & 94.1 & 95.3 & 23.5 & 37.4 & 57.0 & 50.1 & 55.9 & 68.7 & 75.1\\
        HAMNet \cite{li2020multi} & AAAI-20 & 74.5 & 93.3 & 94.3 & 27.1 & 42.3 & 61.6 & 45.6 & 48.5 & 63.1 & 68.8\\
        CCNet \cite{zheng2023cross} & INFFUS-23 & 77.2 & 96.3 & \cellcolor{secondblue}97.2 & 36.4 & 55.2 & \cellcolor{secondblue}72.4 & 50.3 & 52.7 & 69.6 & 75.1 \\
        TOP-ReID \cite{wang2024top} & AAAI-24 & 81.2 & 96.4 & 96.9 & 35.9 & 44.6 & -- & 67.7 & 75.3 & 80.8 & 83.5 \\
        FACENet \cite{zheng2025flare} & INFFUS-25 & 81.5 & 96.9 & -- & 36.2 & 54.1 & -- & \cellcolor{secondblue}69.8 & 77.0 & 81.0 & \cellcolor{secondblue}84.2 \\
        EDITOR \cite{zhang2024magic} & CVPR-24 & 82.1 & 96.4 & 96.9 & 39.0 & 49.3 & -- & 65.6 & 73.8 & 80.0 & 82.3 \\
        MambaPro \cite{wang2025mambapro} & AAAI-25 & 83.9 & 94.7 & 94.9 & 47.0 & 56.5 & -- & 69.5 & 76.9 & 80.6 & 83.8 \\
        PromptMA \cite{zhang2025prompt} & TIP-25 & 85.3 & 97.4 & -- & \cellcolor{secondblue}55.2 & 64.5 & -- & -- & -- & -- & -- \\
        DeMo \cite{wang2025decoupled} & AAAI-25 & 86.2 & \cellcolor{secondblue}97.6 & -- & 49.2 & 59.8 & -- & 68.8 & \cellcolor{secondblue}77.2 & \cellcolor{secondblue}81.5 & 83.8 \\
        Signal \cite{liu2026signal} & AAAI-26 & 86.3 & \cellcolor{secondblue}97.6 & -- & 53.6 & \cellcolor{secondblue}71.9 & -- & -- & -- & -- & -- \\
        $\text{IDEA}^{\dagger}$ \cite{wang2025idea} & CVPR-25 & \cellcolor{secondblue}87.2 & 96.5 & -- & 47.0 & 62.4 & -- & -- & -- & -- & -- \\
        $\text{MODAL}^{\dagger}$ & Ours & \cellcolor{bestblue}\textbf{87.7} & \cellcolor{bestblue}\textbf{98.6} & \cellcolor{bestblue}\textbf{99.0} & \cellcolor{bestblue}\textbf{57.7} & \cellcolor{bestblue}\textbf{73.9} & \cellcolor{bestblue}\textbf{84.3} & \cellcolor{bestblue}\textbf{71.2} & \cellcolor{bestblue}\textbf{79.9} & \cellcolor{bestblue}\textbf{82.9} & \cellcolor{bestblue}\textbf{85.8} \\
        \noalign{\hrule height 1pt}
    \end{tabular}
    }
\end{table*}

\subsection{Comparison with State-of-the-Art Methods}
\noindent \textbf{Multi-modal Re-ID for Pedestrian.}
We compare the proposed method with representative multi-modal pedestrian Re-ID approaches. Specifically, the CNN-based methods include HAMNet~\cite{li2020multi}, PFNet~\cite{zheng2021robust}, IEEE~\cite{wang2022interact}, LRMM~\cite{wu2025lrmm}, and TIENet~\cite{yang2025tienet}. The ViT-based approaches comprise EDITOR~\cite{zhang2024magic}, WTSF-ReID~\cite{yu2025wtsf}, HTT~\cite{wang2024heterogeneous}, and TOP-ReID\cite{wang2024top}. For CLIP-based frameworks, we compare with ICPL-ReID~\cite{li2026icpl}, PromptMA~\cite{zhang2025prompt}, MambaPro~\cite{wang2025mambapro}, DeMo~\cite{wang2025decoupled}, Signal~\cite{liu2026signal}, and IDEA~\cite{wang2025idea}. Notably, IDEA leverages additional generated textual prompts to enhance multi-modal representation learning.
Table \ref{tab.SOTA pedestrian} presents the performance of different multi-modal Re-ID methods on the RGBNT201 dataset. MODAL demonstrates strong effectiveness and superior accuracy by leveraging explicit modal decoupling and irrelevant information suppression. Specifically, MODAL achieves a mAP of $83.8\%$ and a Rank-1 accuracy of $86.1\%$, surpassing the performance of MambaPro by $4.9\%$ in mAP and $2.7\%$ in Rank-1. Compared with DeMo, which employs heuristically decoupling, MODAL outperforms it by $4.8\%$ in mAP and $3.8\%$ in Rank-1. Additionally, when compared with IDEA, which also incorporates text annotations, MODAL achieves improvements of $3.6\%$ in mAP and $4.0\%$ in Rank-1.

\noindent \textbf{Multi-modal Re-ID for Vehicle.}
We further evaluate the performance of comparison methods on three vehicle Multi-modal Re-ID datasets. The comparisons cover CNN-based methods, including IEEE~\cite{wang2022interact}, PFNet~\cite{zheng2021robust}, HAMNet~\cite{li2020multi}, and CCNet~\cite{zheng2023cross}; transformer-based frameworks such as TOP-ReID~\cite{wang2024top}, FACENet~\cite{zheng2025flare}, and EDITOR~\cite{zhang2024magic}; CLIP-driven models, including MambaPro~\cite{wang2025mambapro}, PromptMA~\cite{zhang2025prompt}, DeMo~\cite{wang2025decoupled}, Signal~\cite{liu2026signal}, and IDEA~\cite{wang2025idea}.
IDEA similarly incorporates additional generated textual prompts.
As summarized in Table \ref{tab.SOTA vehicle}. Specifically, MODAL achieves a mAP of 87.7\% and Rank-1 accuracy of 98.6\% on RGBNT100, a mAP of 55.9\% with Rank-1 accuracy of 70.4\% on MSVR310, and a mAP of 71.2\% with Rank-1 accuracy of 79.9\% on WMVEID863. MODAL consistently outperforms both DeMo and IDEA. Notably, on the challenging MSVR310 dataset, MODAL surpasses IDEA by 8.9\% in mAP and 8.0\% in Rank-1 accuracy. These results further demonstrate the effectiveness and generalizability of MODAL in vehicle-based multi-modal Re-ID tasks.

\begin{table*}[t]
    \caption{Performance of missing-modality settings on RGBNT201.($*$ indicates that the MASA is not applied; instead, fully decoupled features are concatenated as the final representation.) “M (X)” means missing the X image modality.}
    \label{tab:missing-modality ReID}
    \centering
    \renewcommand\arraystretch{1.0}
    \setlength\tabcolsep{5pt}
    \resizebox{1\textwidth}{!}
    {
    \begin{tabular}{cc c cccccccccccccc}
        \noalign{\hrule height 1pt}
    &\multicolumn{1}{c}{\multirow{2}{*}{\textbf{Methods}}} & \multicolumn{1}{c}{\multirow{2}{*}{\textbf{Venue}}} &  \multicolumn{2}{c}{\textbf{M (RGB)}} & \multicolumn{2}{c}{\textbf{M (NIR)}} & \multicolumn{2}{c}{\textbf{M (TIR)}} & \multicolumn{2}{c}{\textbf{M (RGB+NIR)}} & \multicolumn{2}{c}{\textbf{M (RGB+TIR)}} & \multicolumn{2}{c}{\textbf{M (NIR+TIR)}} & \multicolumn{2}{c}{\textbf{Average}} \\
    \cmidrule(r){4-5} \cmidrule(r){6-7} \cmidrule(r){8-9} \cmidrule(r){10-11} \cmidrule(r){12-13} \cmidrule(r){14-15} \cmidrule(r){16-17}
        & & & \textbf{\emph{m}AP} & \textbf{R-1}  & \textbf{\emph{m}AP} & \textbf{R-1} & \textbf{\emph{m}AP} & \textbf{R-1} & \textbf{\emph{m}AP} & \textbf{R-1} & \textbf{\emph{m}AP} & \textbf{R-1} & \textbf{\emph{m}AP} & \textbf{R-1} & \textbf{\emph{m}AP} & \textbf{R-1} \\
    \hline
    & HACNN \cite{Li2018HarmoniousAN} & CVPR-18 & 12.5 & 11.1 & 20.5 & 19.4 & 16.7 & 13.3 & 9.2 & 6.2 & 6.3 & 2.2 & 14.8 & 12.0 & 13.3 & 10.7 \\
    & MLFN \cite{Chang2018MultilevelFN} & CVPR-18 & 20.2 & 18.9 & 21.1 & 19.7 & 17.6 & 11.1 & 13.2 & 12.1 & 8.3 & 3.5 & 13.1 & 9.1 & 15.6 & 12.4 \\
    & PCB \cite{Sun2017BeyondPM} & ECCV-18 & 23.6 & 24.2 & 24.4 & 25.1 & 19.9 & 14.7 & 20.6 & 23.6 & 11.0 & 6.8 & 18.6 & 14.4 & 19.7 & 18.1 \\
    & OSNet \cite{Zhou2019OmniScaleFL} & ICCV-19 & 19.8 & 17.3 & 21.0 & 19.0 & 18.7 & 14.6 & 12.3 & 10.9 & 9.4 & 5.4 & 13.0 & 10.2 & 15.7 & 12.9 \\
    & PFNet \cite{zheng2021robust} & AAAI-21 & - & - & 31.9 & 29.8 & 25.5 & 25.8 & - & - & - & - & 26.4 & 23.4 & - & - \\
    & TOP-ReID \cite{wang2024top} & AAAI-24 & 54.4 & 57.5 & 64.3 & 67.6 & 51.9 & 54.5 & 35.3 & 35.4 & 26.2 & 26.0 & 34.1 & 31.7 & 44.4 & 45.4 \\
    & MFRNet \cite{Feng2025MultiModalOR} & ICML-25 & 64.7 & 65.2 & 72.3 & 76.1 & 51.6 & 49.5 & 41.4 & 43.4 & \cellcolor{secondblue}27.3 & \cellcolor{bestblue}\textbf{27.9} & 37.2 & 35.6 & 49.1 & 49.6 \\
    & IDEA \cite{wang2025idea} & CVPR-25 & 62.9 & -- & 71.5 & -- & \cellcolor{secondblue}58.4 & -- & 43.3 & -- & 27.1 & -- & 39.9 & -- & 50.5 & -- \\
    & DeMo \cite{wang2025decoupled} & AAAI-25 & 63.3 & 65.3 & 72.6 & 75.7 & 56.2 & 54.1 & \cellcolor{secondblue}45.6 & \cellcolor{secondblue}46.5 & 26.3 & 24.9 & 40.3 & 38.5 & 50.7 & 50.8 \\
    \hline
    & MODAL$^*$ & Ours & \cellcolor{secondblue}67.9 & \cellcolor{secondblue}68.7 & \cellcolor{secondblue}75.3 & \cellcolor{secondblue}76.4 & 58.1 & \cellcolor{secondblue}56.0 & \cellcolor{secondblue}46.8 & 46.3 & 27.1 & 25.6 & \cellcolor{secondblue}41.9 & \cellcolor{secondblue}42.7 & \cellcolor{secondblue}52.9 & \cellcolor{secondblue}52.6 \\
    & MODAL & Ours & \cellcolor{bestblue}\textbf{68.6} & \cellcolor{bestblue}\textbf{70.3} & \cellcolor{bestblue}\textbf{75.8} & \cellcolor{bestblue}\textbf{78.1} & \cellcolor{bestblue}\textbf{59.2} & \cellcolor{bestblue}\textbf{59.8} & \cellcolor{bestblue}\textbf{50.4} & \cellcolor{bestblue}\textbf{51.1} & \cellcolor{bestblue}\textbf{28.8} & \cellcolor{secondblue}26.3 & \cellcolor{bestblue}\textbf{42.4} & \cellcolor{bestblue}\textbf{43.9} & \cellcolor{bestblue}\textbf{54.2} & \cellcolor{bestblue}\textbf{54.9} \\
        \noalign{\hrule height 1pt}
    \end{tabular}
    }
    \vspace{-1.5em}
\end{table*}

\begin{table*}[t]
    \caption{Performance of modality-mismatched Scenarios.($*$ indicates that the MASA is not applied; instead, fully decoupled features are concatenated as the final representation.)}
    \label{tab: mismatch ReID}
    \centering
    \renewcommand\arraystretch{1.1}
    \setlength\tabcolsep{5pt}
    \resizebox{0.72\textwidth}{!}
    {
    \begin{tabular}{cc c cc cc cc cc cc}
        \noalign{\hrule height 1pt}
    &\multicolumn{1}{c}{\multirow{2}{*}{\textbf{Methods}}} & \multicolumn{1}{c}{\multirow{2}{*}{\textbf{Venue}}} & \multicolumn{2}{c}{\textbf{RT-to-NT}} & \multicolumn{2}{c}{\textbf{RT-to-N}} & \multicolumn{2}{c}{\textbf{R-to-N}} & \multicolumn{2}{c}{\textbf{R-to-NT}} & \multicolumn{2}{c}{\textbf{Average}} \\
    \cmidrule(r){4-5} \cmidrule(r){6-7} \cmidrule(r){8-9} \cmidrule(r){10-11} \cmidrule(r){12-13}
    & & & \textbf{\emph{m}AP} & \textbf{R-1} & \textbf{\emph{m}AP} & \textbf{R-1} & \textbf{\emph{m}AP} & \textbf{R-1} & \textbf{\emph{m}AP} & \textbf{R-1} & \textbf{\emph{m}AP} & \textbf{R-1} \\
    \hline
    &EDITOR \cite{zhang2024magic} & CVPR-24  & 27.3 & 27.9 & 2.80 & 0.0 & 4.0 & 2.3 & 4.3 & 4.1 & 9.6 & 8.6 \\
    &TOP-ReID \cite{wang2024top} & AAAI-24 & \cellcolor{secondblue}43.0 & \cellcolor{secondblue}44.6 & \cellcolor{secondblue}14.4 & \cellcolor{bestblue}\textbf{13.3} & \cellcolor{bestblue}\textbf{15.4} & \cellcolor{bestblue}\textbf{14.0} & \cellcolor{secondblue}11.9 & \cellcolor{bestblue}\textbf{8.7} & \cellcolor{secondblue}21.2 & \cellcolor{secondblue}20.2 \\
    \hline
    &MODAL$^*$ & Ours & 37.8 & 32.9 & 5.2 & 4.9 & 5.3 & 4.3 & 4.9 & 1.8 & 13.3 & 11.0 \\
    &MODAL & Ours & \cellcolor{bestblue}\textbf{54.9} & \cellcolor{bestblue}\textbf{52.9} & \cellcolor{bestblue}\textbf{17.5} & \cellcolor{secondblue}13.2 & \cellcolor{secondblue}13.7 & \cellcolor{secondblue}8.5 & \cellcolor{bestblue}\textbf{12.9} & \cellcolor{secondblue}6.9 & \cellcolor{bestblue}\textbf{24.8} & \cellcolor{bestblue}\textbf{20.4} \\
        \noalign{\hrule height 1pt}
    \end{tabular}
    }\vspace{-1em}
\end{table*}

\begin{figure}[t]
    \centering
    \includegraphics[width=1\columnwidth]{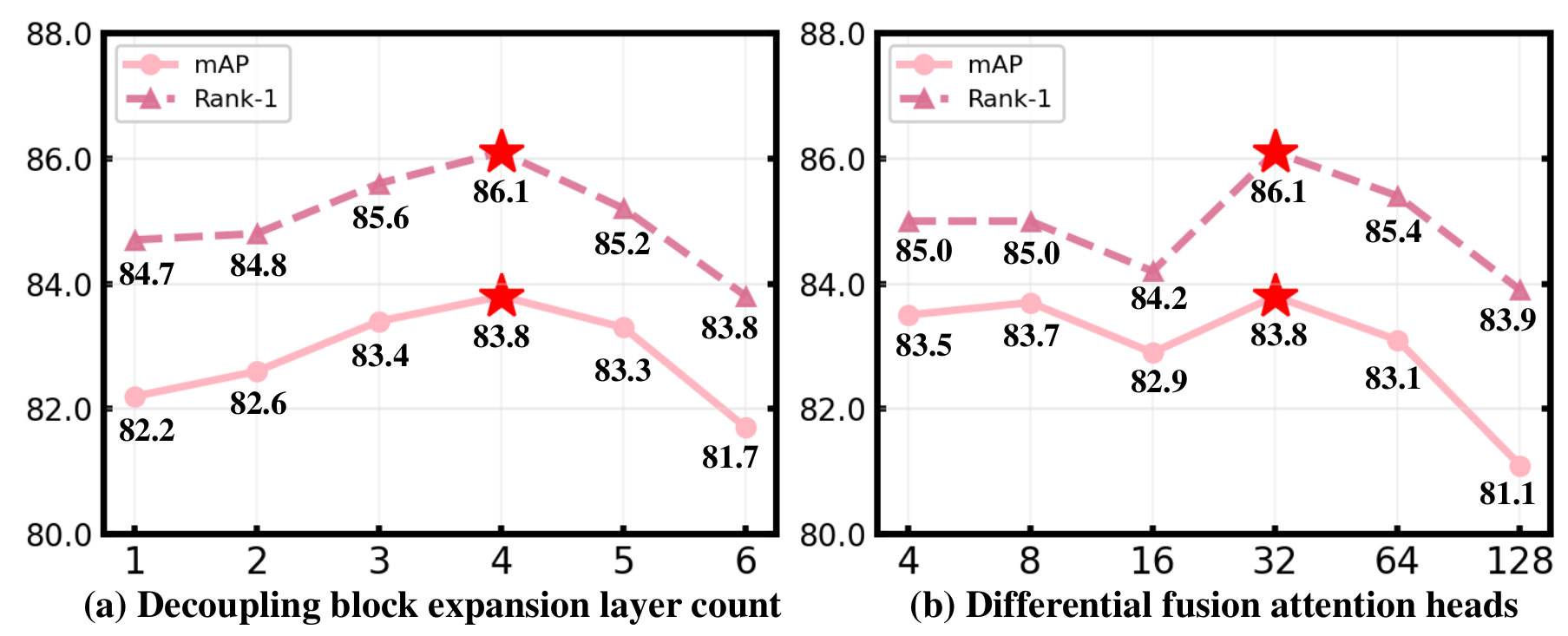}
    \caption{Impact of the MFSD unfolding depth and the number of heads in TIDF blocks on model performance.}
    \label{fig. visual line plot}
    \vspace{-6pt} 
\end{figure}

\subsection{Evaluation under Incomplete and Mismatched Modalities}
\label{sec:Miss modal}
In real-world multi-modal Re-ID deployments, sensor failures, occlusions, and asynchronous acquisition often lead to incomplete inputs or modality inconsistencies between query and gallery samples. Since most existing methods implicitly assume complete modality availability, their performance tends to degrade substantially under such conditions. To rigorously assess robustness, we conduct additional evaluations under modality-missing and modality-mismatched protocols.
\noindent \textbf{Modality Missing Protocol.} 
Following prior works~\cite{zhang2024magic,wang2025mambapro}, modality missing is simulated at test time by removing one or multiple image modalities while keeping the training setting unchanged. On RGBNT201, we consider six configurations: three single-modality missing cases (M(RGB), M(NIR), M(TIR)) and three dual-modality missing cases (M(RGB+NIR), M(RGB+TIR), M(NIR+TIR)). During inference, the corresponding modality inputs are discarded, and all compared methods are evaluated without additional retraining. Performance is measured using mAP and Rank-1 accuracy.

The results are reported in Table~\ref{tab:missing-modality ReID}. MODAL consistently achieves the highest performance across all missing-modality configurations. 
Specifically, when averaging the performance over the six missing-modality configurations, MODAL surpasses IDEA by 3.7\% in mAP. Compared with DeMo, MODAL achieves a 3.5\% improvement in average mAP and a 4.1\% gain in average Rank-1 accuracy.
This robustness stems from the structured disentanglement of modality-specific and shared components, which enables the proposed Modality-Aware Subspace Activation (MASA) to restrict inference to valid shared subspaces when certain modalities are absent. As a result, matching is performed on structurally consistent representations rather than implicitly entangled features, substantially alleviating performance degradation under severe modality incompleteness.

\noindent \textbf{Modality Mismatch Protocol.} 
We further evaluate modality-mismatched scenarios, where the modality sets of query and gallery samples differ. Specifically, we consider several cross-modality retrieval settings on RGBNT201, including RT-to-NT, RT-to-N, R-to-N, and R-to-NT, reflecting practical cross-sensor deployment conditions. No additional alignment or retraining is performed during inference.
The quantitative comparisons are summarized in Table~\ref{tab: mismatch ReID}.
EDITOR achieves only 9.6\% average mAP and 8.6\% average Rank-1 across the four mismatched settings, demonstrating that conventional complete-modal methods suffers severe degradation under modality inconsistency.  Although TOP-ReID alleviates this issue by reconstructing missing-modal information, MODAL still surpasses it by 3.6\% in average mAP and 0.2\% in average Rank-1. Notably, this advantage is achieved without relying on any reconstruction strategy, but through principled disentanglement and adaptive cross-selection.  In modality-mismatched scenarios, where query and gallery modalities are inherently misaligned, MASA dynamically selects mutually supported shared components across samples, preventing similarity estimation from being dominated by inconsistent modality-specific signals.  This selective alignment mechanism contributes to more reliable cross-modal identity matching under severe modality discrepancy.

\subsection{Ablation Studies}

\noindent \textbf{Effectiveness of Multi-modal Feature Sparse Decoupling.} Table~\ref{tab. ablation MFSD} compares different decoupling strategies applied jointly to both visual and textual features, including no-decoupling, linear mapping, HDM~\cite{wang2025decoupled}, and the proposed MFSD. Let $K$ denote the count of Uni-modal, Bi-modal, and Tri-modal blocks. Notably, HDM was originally designed for visual features only, we extend it here for a fair multi-modal comparison (both visual and textual). The proposed MFSD module achieves the most performance gains, surpassing HDM by 3.9\% mAP and 2.5\% Rank-1. Moreover, $\text{MFSD}_{K=1}$ attains higher accuracy than HDM with fewer parameters (12.5M vs. 14.7M), highlighting the superior efficiency of our decoupling design. As shown in Fig.~\ref{fig. visual line plot}(a), increasing $K$ further improves the decoupled features.

\noindent \textbf{Effectiveness of Text-Image Differential Filtering.} 
Table~\ref{tab. ablation TIDF} compares different text-image fusion strategies, demonstrating the effectiveness and efficiency of our proposed TIDF module. Compared with standard cross-attention fusion, The proposed TIDF achieves notable improvements of 3.7\% mAP and 3.9\% Rank-1 while using fewer parameters (8.5M vs. 10.4M), highlighting its superior fusion efficiency. We also compare with the InverseNet~\cite{wang2025idea} (combined with cross-attention) and observe that MODAL still achieves an additional 2.2\% mAP and 3.2\% Rank-1 improvement. Furthermore, we investigate the effect of varying the number of heads in TIDF blocks, as illustrated in Fig.~\ref{fig. visual line plot}(b). The results show that using $32$ heads achieves the best performance, as the induced sparsity enables the network to capture diverse information across fewer dimensions.

\begin{table}[t]
\centering
\caption{Ablation study of feature decoupling methods on RGBNT201. ($K$ denotes the number of unfolding blocks.)}
\label{tab. ablation MFSD}
\resizebox{0.44\textwidth}{!}{
\begin{tabular}{c|cccc}
\toprule
\multirow{2}{*}{\textbf{Methods}} & \multicolumn{3}{c}{\textbf{Metric}} & \multirow{2}{*}{\textbf{Params}} \\
\cmidrule(lr){2-4}
 & \textbf{mAP} & \textbf{Rank-1} & \textbf{Rank-5} & \\
\midrule
No-decoupling & \cellcolor{abred5}77.2 & \cellcolor{abred5}79.8 & \cellcolor{abred5}85.8 & 0 \\
Linear Mapping & \cellcolor{abred4}78.1 & \cellcolor{abred4}81.1 & \cellcolor{abred3}88.6 & 6.3M \\
HDM~\cite{wang2025decoupled} & \cellcolor{abred3}79.9 & \cellcolor{abred3}83.6 & \cellcolor{abred4}85.9 & 14.7M \\
$\text{MFSD}_{K=1}$ (ours) & \cellcolor{abred2}82.2 & \cellcolor{abred2}84.7 & \cellcolor{abred1}\textbf{91.1} & 12.5M \\
$\text{MFSD}_{K=4}$ (ours) & \cellcolor{abred1}\textbf{83.8} & \cellcolor{abred1}\textbf{86.1} & \cellcolor{abred2}90.3 & 34.9M \\
\noalign{\hrule height 1pt}
\end{tabular}
}
\end{table}

\begin{table}[t]
\centering
\caption{Ablation study on text-image fusion methods on RGBNT201.}
\label{tab. ablation TIDF}
\resizebox{0.42\textwidth}{!}{
\begin{tabular}{c|cccc}
\toprule
\multirow{2}{*}{\textbf{Methods}} & \multicolumn{3}{c}{\textbf{Metric}} & \multirow{2}{*}{\textbf{Params}} \\
\cmidrule(lr){2-4} 
 & \textbf{mAP} & \textbf{Rank-1} & \textbf{Rank-5} & \\
\midrule
Concatenation & \cellcolor{abred4}75.1 & \cellcolor{abred4}76.8 & \cellcolor{abred4}84.7 & 0 \\
Cross Attention & \cellcolor{abred3}80.1 & \cellcolor{abred3}82.2 & \cellcolor{abred2}88.9 & 10.4M \\
InverseNet~\cite{wang2025idea} & \cellcolor{abred2}81.6 & \cellcolor{abred2}82.4 & \cellcolor{abred3}88.2 & 13.0M \\
TIDF (ours) & \cellcolor{abred1}\textbf{83.8} & \cellcolor{abred1}\textbf{86.1} & \cellcolor{abred1}\textbf{90.3} & 8.5M \\
\noalign{\hrule height 1pt}
\end{tabular}
}
\end{table}

\begin{table}[t]
\centering
\caption{Ablation study on different structural designs of MFSD, evaluated on RGBNT201 dataset.}
\label{tab: struct of MFSD}
\resizebox{0.40\textwidth}{!}{
\begin{tabular}{l|ccc}
\toprule
\multirow{2}{*}{\textbf{Learnable Transform}} & \multicolumn{3}{c}{\textbf{RGBNT201}} \\
\cmidrule(lr){2-4}  
 & \textbf{mAP} & \textbf{Rank1} & \textbf{Rank5} \\
\midrule
None (only bottleneck) & \cellcolor{abred5}80.8 & \cellcolor{abred5}82.1 & \cellcolor{abred5}89.0 \\
Single Linear Layer & \cellcolor{abred3}82.4 & \cellcolor{abred2}84.5 & \cellcolor{abred3}90.1 \\
Group convolution & \cellcolor{abred2}82.7 & \cellcolor{abred3}85.0 & \cellcolor{abred4}89.2 \\
Mlp (Expansion=2) & \cellcolor{abred4}81.8 & \cellcolor{abred4}83.5 & \cellcolor{abred1}\textbf{90.6} \\
Linear layers w/ dropout & \cellcolor{abred1}\textbf{83.8} & \cellcolor{abred1}\textbf{86.1} & \cellcolor{abred2}90.3 \\
\noalign{\hrule height 1pt}
\end{tabular}
}
\vspace{-6pt} 
\end{table}

\begin{table}[t]
\centering
\caption{Ablation study on text and image usage on RGBNT201.}
\label{tab. ablation text and image usage}
\resizebox{0.34\textwidth}{!}{
\begin{tabular}{c|ccc}
\toprule
\multirow{2}{*}{\textbf{Data Usage}} & \multicolumn{3}{c}{\textbf{Metric}} \\
\cmidrule(lr){2-4}
& \textbf{mAP} & \textbf{Rank-1} & \textbf{Rank-5} \\
\midrule
Text Only & \cellcolor{abred5}10.2 & \cellcolor{abred5}13.9 & \cellcolor{abred5}20.1 \\
Image Only & \cellcolor{abred3}81.4 & \cellcolor{abred3}82.9 & \cellcolor{abred3}87.6 \\
Text \& Image & \cellcolor{abred1}\textbf{83.8} & \cellcolor{abred1}\textbf{86.1} & \cellcolor{abred1}\textbf{90.3} \\
\noalign{\hrule height 1pt}
\end{tabular}
}
\end{table}

\begin{table}[t]
\centering
\caption{Ablation study on text granularity on RGBNT201.}
\label{tab. ablation text granularity}
\resizebox{0.4\textwidth}{!}{
\begin{tabular}{c|ccc}
\toprule
\multirow{2}{*}{\textbf{Text Granularity}} & \multicolumn{3}{c}{\textbf{Metric}} \\
\cmidrule(lr){2-4}
& \textbf{mAP} & \textbf{Rank-1} & \textbf{Rank-5} \\
\midrule
w/o Text & \cellcolor{abred4}81.4 & \cellcolor{abred4}82.9 & \cellcolor{abred4}87.6 \\
w/ Learnable Prompt & \cellcolor{abred3}81.9 & \cellcolor{abred2}85.8 & \cellcolor{abred3}90.1 \\
w/ Fixed Caption & \cellcolor{abred2}82.1 & \cellcolor{abred3}84.7 & \cellcolor{abred1}\textbf{91.4} \\
w/ Prompt \& Caption & \cellcolor{abred1}\textbf{83.8} & \cellcolor{abred1}\textbf{86.1} & \cellcolor{abred2}90.3 \\
\noalign{\hrule height 1pt}
\end{tabular}
}
\end{table}

\noindent \textbf{Different Structural Designs of MFSD}
To explore the most effective architectural form of the learnable transformations $\phi(\cdot)$ and $\psi(\cdot)$ in the UFD, BFD, and TFD blocks, we compare our adopted implementation with several alternative structures. Specifically, we conduct ablation experiments by replacing the adopted design with several variants, including retaining only the bottleneck layer without additional learnable transformations, using a single linear layer, using a group convolutional layer, and using an MLP with an expansion factor of 2. The results on RGBNT201 are reported in Table~\ref{tab: struct of MFSD}. Among all configurations, the bottleneck-only variant yields the weakest performance, indicating that stronger transformation capacity is beneficial for effective sparse feature decoupling. Introducing a single linear layer brings moderate improvement, while group convolution, which is related to the strategy adopted in LCSC~\cite{sreter2018learned}, further improves the results. Overall, our design based on linear layers and dropout-based nonlinear activations achieves the best performance, outperforming the bottleneck-only variant by 3.0\% in mAP and 4.0\% in Rank-1, demonstrating that it provides the most effective structural choice among the compared alternatives.

\noindent \textbf{Ablations of Text Granularity.} 
We further investigate the effect of text granularity in the proposed MODAL framework. As shown in Table~\ref{tab. ablation text and image usage}, text-only input yields poor performance (10.2\% mAP), confirming that the generated text features are coarse-grained and lack discriminative power. The proposed method exploits its semantic cues through a differential filtering strategy, providing an effective way to utilize in-discriminative text information. Table~\ref{tab. ablation text granularity} further compares different text granularity configurations, where prompt denotes learnable text embeddings and caption represents MLLM-generated textual descriptions. The best performance is achieved when combining learnable prompts with MLLM-generated captions, validating that the TIDF module effectively leverages coarse-grained textual semantics to suppress task-irrelevant visual features.

\begin{figure}[t]
    \centering
    \includegraphics[width=1\columnwidth]{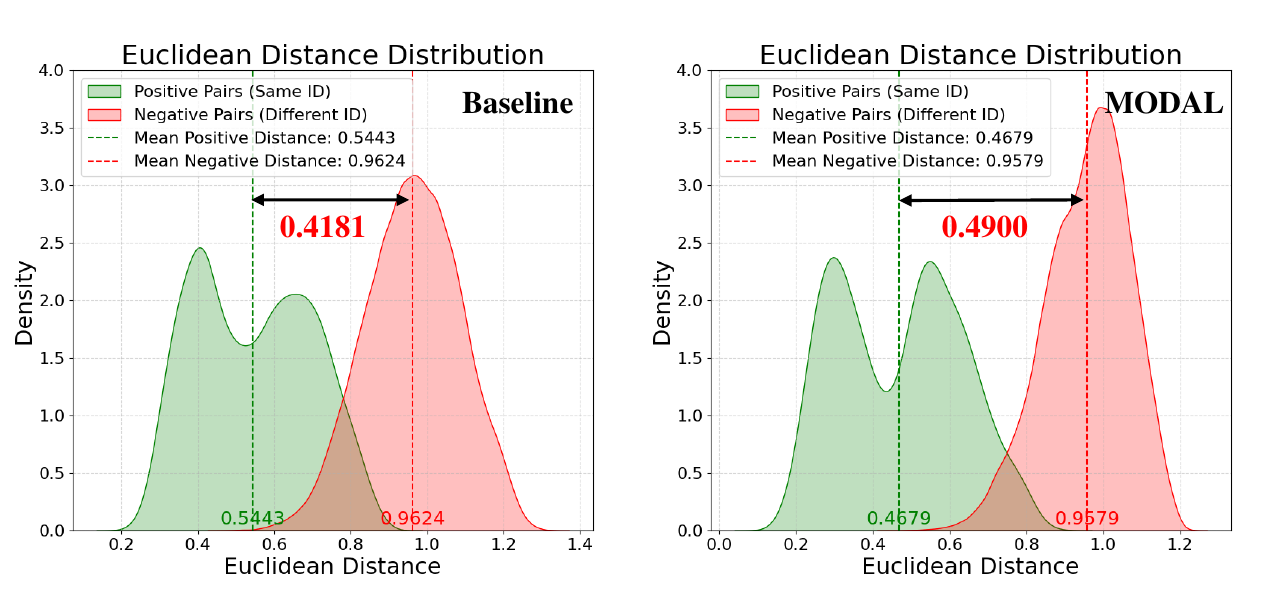}
    \caption{Visualization comparison of the Euclidean distance distributions between MODAL and the baseline method.}
    \label{fig. visual euc similarity}
    \vspace{-6pt} 
\end{figure}

\begin{figure}[t]
    \centering
    \includegraphics[width=0.99  \columnwidth]{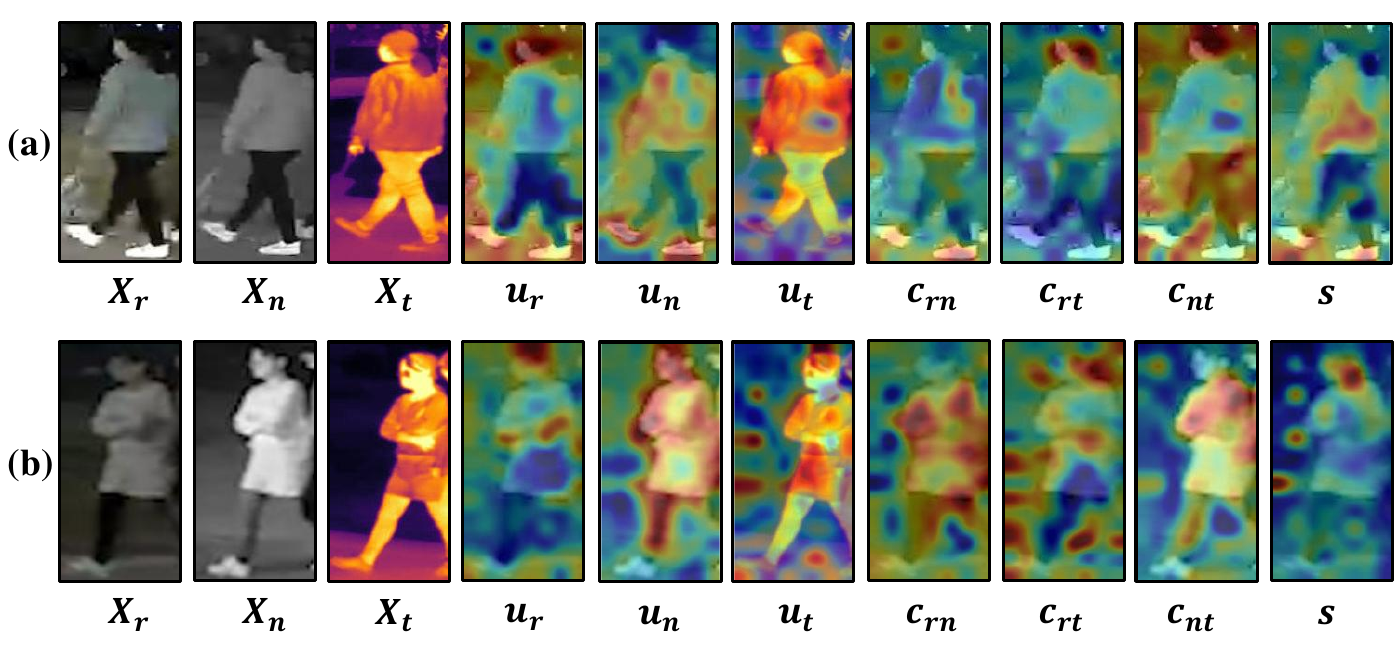}
    \caption{Visualization of attention maps for the decoupled uni-modal ($\bm{h}_r$, $\bm{h}_n$, $\bm{h}_t$), bi-modal ($\bm{h}_{rn}$, $\bm{h}_{rt}$, $\bm{h}_{nt}$), and tri-modal (${h}_{rnt}$) feature components produced by MFSD. Diffdifferent sparse components attend to distinct object regions and exhibit limited overlap, indicating effective disentanglement of modality-specific and shared representations.}
    \label{fig. visual activation MFSD}
    \vspace{-6pt} 
\end{figure}

\subsection{Visualization Analysis}

\noindent \textbf{Euclidean Distance Distribution.} 
Fig.~\ref{fig. visual euc similarity} presents the Euclidean distance distributions of test features extracted by the baseline model (without the proposed MFSD and TIDF modules) and the complete MODAL. Compared with the baseline, the proposed MODAL yields a noticeably larger average distance gap between positive and negative pairs (0.4900 vs. 0.4181), indicating that the proposed MFSD and TIDF effectively enhanced feature separability.

\noindent \textbf{Attention Maps with Decoupled Features.} 
Fig.~\ref{fig. visual activation MFSD} illustrates the attention responses within the uni-modal, bi-modal, and tri-modal representations. 
For instance on Fig.~\ref{fig. visual activation MFSD}(a), no activation is observed for pedestrian pants in $\bm{h}_{r}$, $\bm{h}_{rn}$, and ${h}_{rnt}$, suggesting that RGB-related patterns disregard black pants under low-light conditions. Moreover, the attention area of the pedestrian's upper body in ${h}_{rnt}$ shows minimal overlap with those of other decoupled features, indicating that the features have been successfully disentangled into distinct components. The results demonstrate that MODAL not only captures diverse aspects of the object, but also offers a degree of interpretability. These findings validate that our explicit decoupling strategy effectively mitigates pattern conflicts.

\begin{figure}[t]
    \centering
    \includegraphics[width=0.98\columnwidth]{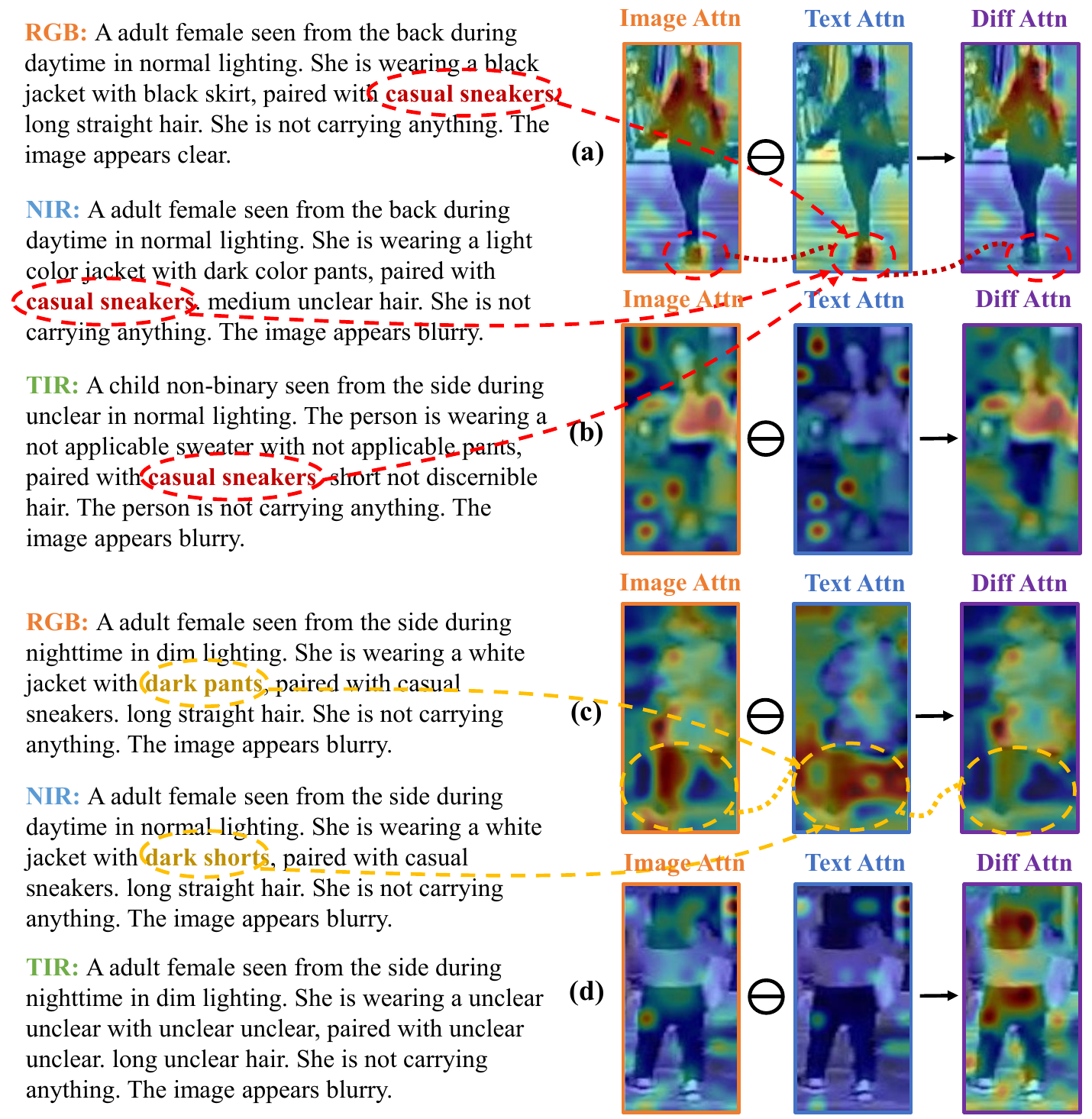}
    \caption{Visualization of attention maps for the proposed Text-Image Differential Filtering (TIDF) module. The left panel presents the RGB, NIR, and TIR textual descriptions of two representative instances, where salient textual attributes are highlighted and linked to their corresponding textual attention regions. The right panel shows, for four examples, the image attention, text attention, and differential attention maps, respectively. }
    \label{fig. visual activation TIDF}
    \vspace{-6pt} 
\end{figure}

\noindent \textbf{Attention Maps for Text-Image Differential Filtering.} 
We visualize the attention distributions of the image branch, the text branch, and their differential attention maps. As illustrated in Fig.~\ref{fig. visual activation TIDF}, the differential filtering operation effectively removes task irrelevant or misleading attention patterns while reinforcing task-relevant cues. For instance, in Fig.~\ref{fig. visual activation TIDF} (a), the text primarily focuses on the pedestrian’s foot region, after differential filtering, such task irrelevant attention is suppressed. Similarly, Fig.~\ref{fig. visual activation TIDF} (b) and Fig.~\ref{fig. visual activation TIDF} (c) show that excessive responses to background regions and the lower body are mitigated. In contrast, Fig.~\ref{fig. visual activation TIDF} (d) demonstrate that textual guidance helps the model reallocate attention toward the true target regions, thereby enhancing the expression of task relevant features.

\section{Conclusions}
In this paper, we present MODAL, a principled and transparent framework for multi-modal object Re-ID. The proposed Multi-modal Feature Sparse Decoupling module performs structured deep disentanglement to explicitly decompose uni-modal specific, bi-modal shared, and tri-modal shared representations, establishing a well-defined subspace foundation for cross-modal interaction. Built upon this decomposition, the Modality-Aware Subspace Activation adaptively activates structurally consistent shared components during inference, enabling reliable matching under arbitrary modality combinations, including incomplete and mismatched scenarios. In addition, the Text-Image Differential Filtering module leverages semantic guidance from textual features to suppress task-irrelevant or disruptive visual responses, further enhancing discriminative consistency. By jointly integrating principled disentanglement and adaptive cross-selection, MODAL achieves robust multi-modal alignment while maintaining structural interpretability and practical generality. Extensive experiments on four benchmark datasets verify its effectiveness under both standard, modal miss and modal missmatch evaluation settings.

\section{Future Work}
Overall, the proposed framework demonstrates consistent effectiveness across standard and challenging evaluation protocols, validating the benefits of structured disentanglement and text-guided differential filtering. Quantitative ablations confirm that textual cues help suppress task-irrelevant responses and enhance discriminative representation learning. Nevertheless, visualization analysis suggests that the suppression behavior is not always strictly limited to background regions. In some cases, overly aggressive modulation may also weaken discriminative foreground cues, potentially contributing to occasional failure cases. Moreover, although the proposed MFSD module is reasonably efficient and $\text{MFSD}_{K=1}$ already provides strong performance, the best results are achieved with $K=4$, which inevitably incurs additional computation and slightly reduces inference speed. Although this overhead is not substantial, it should be considered in practical deployment. In addition, while MLLM-generated textual annotations bring measurable gains, their generation cost has not been explicitly evaluated and deserves further study for large-scale or real-time applications. These issues provide meaningful directions for future work toward improving the robustness, efficiency, and practical applicability of the proposed framework.

\ifCLASSOPTIONcaptionsoff
  \newpage
\fi

\bibliographystyle{IEEEtran}
\nocite{*}
\bibliography{IEEEabrv,IEEEexample}

\end{document}